\documentclass[acmsmall]{acmart}

\usepackage{algorithmic}
\usepackage[ruled]{algorithm2e}
\usepackage{graphicx}
\usepackage{subfigure}
\definecolor{revisecolor}{RGB}{0,0,0}
\definecolor{commentcolor}{RGB}{102,153,204}

\usepackage{multirow}

\AtBeginDocument{%
  }

\setcopyright{acmlicensed}
\acmJournal{TOMM}
\acmYear{2023} 
\acmPrice{15.00} \acmVolume{1} \acmNumber{1} \acmArticle{1} \acmMonth{9}
\acmDOI{10.1145/3624015}

\begin{document}

\title{PLACE dropout: A Progressive Layer-wise and Channel-wise Dropout for Domain Generalization}


\author{Jintao Guo}
\orcid{0000-0003-1101-4443}
\affiliation{%
  \institution{The State Key Laboratory for Novel Software Technology, Nanjing University}
  \state{Nanjing}
  \country{China}}
\email{guojintao@smail.nju.edu.cn}

\author{Lei Qi}
\authornote{Corresponding authors: Yinghuan Shi and Lei Qi.}
\orcid{0000-0001-7091-0702}
\affiliation{%
  \institution{The School of Computer Science and Engineering, Southeast University and
  Key Laboratory of New Generation Artificial Intelligence Technology and Its Interdisciplinary Applications (Southeast University), Ministry of Education}
  \state{Nanjing}
  \country{China}}
\email{qilei.cs@gmail.com}

\author{Yinghuan Shi}
\orcid{0009-0008-0722-9744}
\authornotemark[1]
\affiliation{%
\institution{The State Key Laboratory for Novel Software Technology, Nanjing University}
  \state{Nanjing}
  \country{China}}
\email{syh@nju.edu.cn}

\author{Yang Gao}
\orcid{0000-0002-2488-1813}
\affiliation{%
\institution{The State Key Laboratory for Novel Software Technology, Nanjing University}
\state{Nanjing}
  \country{China}}
\email{gaoy@nju.edu.cn}

\renewcommand{\shortauthors}{Guo, et al.}

\begin{abstract}
Domain generalization (DG) aims to learn a generic model from multiple observed source domains that generalizes well to arbitrary unseen target domains without further training. The major challenge in DG is that the model inevitably faces a severe overfitting issue due to the domain gap between source and target domains. To mitigate this problem, some dropout-based methods have been proposed to resist overfitting by discarding part of the representation of the intermediate layers. However, we observe that most of these methods only conduct the dropout operation in some specific layers, leading to an insufficient regularization effect on the model. We argue that applying dropout at multiple layers can produce stronger regularization effects, which could alleviate the overfitting problem on source domains more adequately than previous layer-specific dropout methods. In this paper, we develop a novel layer-wise and channel-wise dropout for DG, which randomly selects one layer and then randomly selects its channels to conduct dropout. Particularly, the proposed method can generate a variety of data variants to better deal with the overfitting issue. We also provide theoretical analysis for our dropout method and prove that it can effectively reduce the generalization error bound. Besides, we leverage the progressive scheme to increase the dropout ratio with the training progress, which can gradually boost the difficulty of training the model to enhance its robustness. Extensive experiments on three standard benchmark datasets have demonstrated that our method outperforms several state-of-the-art DG methods. Our code is available at \textcolor{magenta}{\href{https://github.com/lingeringlight/PLACEdropout}{https://github.com/lingeringlight/PLACEdropout}}.
\end{abstract}





\maketitle

\section{Introduction}

Deep learning has achieved tremendous progress in various tasks over the last few years. 
Under the assumption that training and test data come from similar data distributions, deep convolutional neural networks have shown remarkable ability in a wide range of visual applications \cite{yang2021densely,shi2022shuffle,li2023iomatch}.
However, their trained models often overfit the training data, leading to the inferior performance on out-of-distribution data \cite{li2017deeper}.
Such catastrophic performance degradation caused by distribution discrepancy (\textit{i.e.}, domain shift \cite{pan2009survey}) hinders the applications of deep neural networks, as in reality training and test data are often from different distributions \cite{mahajan2021domain,xu2021fourier}.
To address this issue, domain adaptation (DA) has been proposed to narrow the potential distribution discrepancy by leveraging labeled source domains and unlabeled target domain to jointly train the model \cite{wu2022instance,xu2022towards}.

Unfortunately, despite their success, DA methods could not guarantee the model performance on unknown target domains that have not been seen during training \cite{wang2021learning,shu2021open}.
Since target data cannot always be available in real-world scenarios, collecting data from all potential target domains is expensive and impractical. 
Moreover, even if data from the target domain can be obtained, DA methods need to re-train the model on the new target domain, which is hard to achieve in reality. 
Aware of this fact, domain generalization (DG) has been proposed as a more challenging yet practical problem setting, which aims to train a model with multiple different but related source domains that can generalize well to arbitrary unseen target domains without re-training \cite{zhou2021survey,wang2022generalizing}. 
The DG problem has aroused wide attention from researchers and many existing works have shown promising results by utilizing domain-invariant learning \cite{matsuura2020domain,fan2021adversarially,chen2021style}, meta-learning \cite{balaji2018metareg,dou2019domain,zhang2022more,wang2023generalizable}, data augmentation \cite{volpi2018generalizing,zhou2020deep,wang2022feature}, regularization strategies \cite{shi2020informative,huang2020self,qi2023unsupervised}, \textit{etc.}

\begin{figure}[tbp!]
    \centering
    \subfigure[DeepAll \cite{zhou2020deep}]{
    \begin{minipage}[t]{0.23\linewidth} 
    \centering
    \includegraphics[scale=0.22]{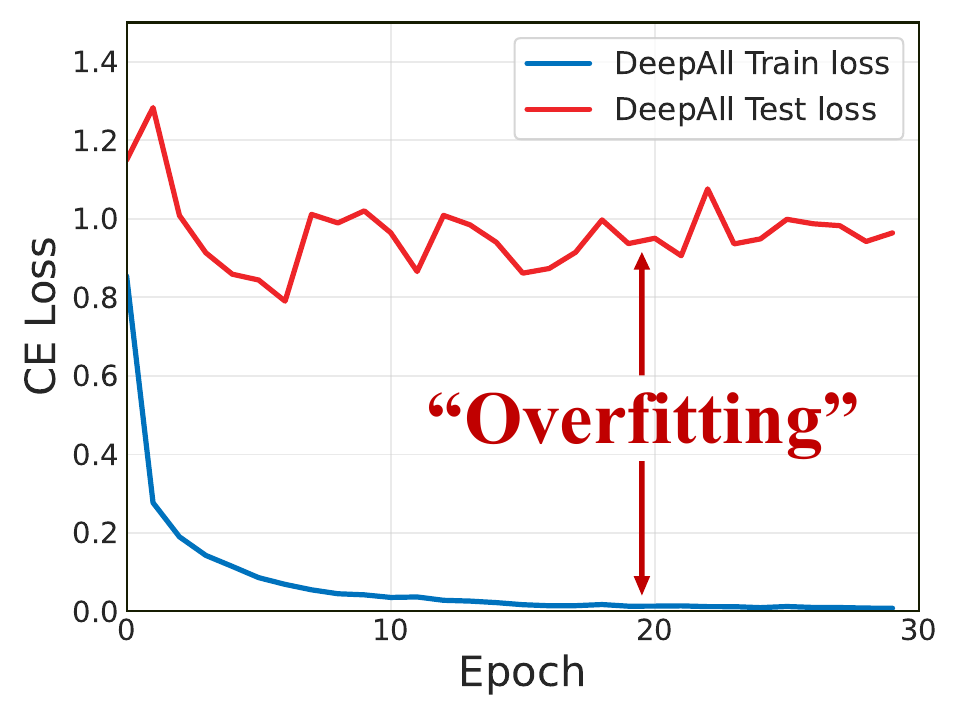} 
    \end{minipage}
    \label{fig:DeepAll CELoss}
    }
    \subfigure[RSC \cite{huang2020self}]{
    \begin{minipage}[t]{0.23\linewidth} 
    \centering
    \includegraphics[scale=0.22]{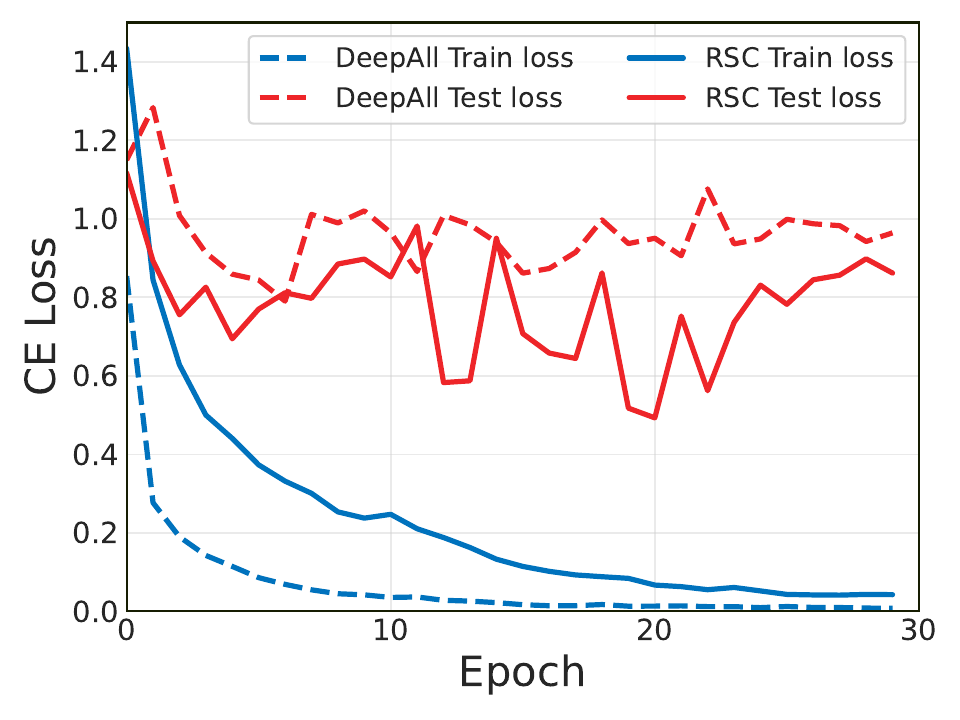}
    \end{minipage}
    \label{fig:RSC CELoss}
    }
    \subfigure[InfoDrop \cite{shi2020informative}]{
    \begin{minipage}[t]{0.23\linewidth} 
    \centering
    \includegraphics[scale=0.22]{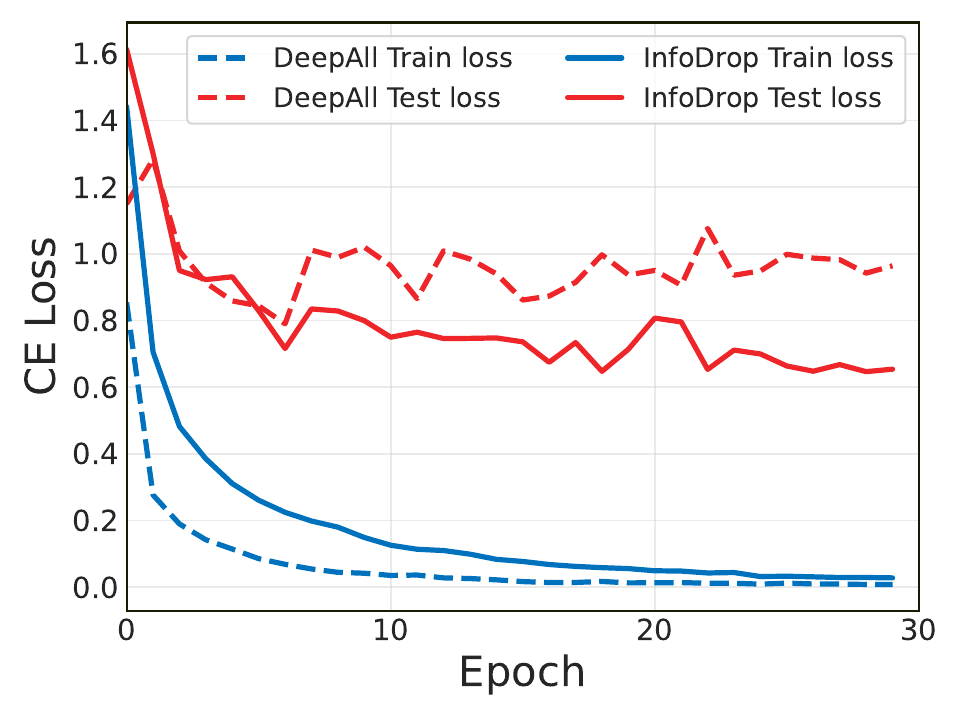}
    \end{minipage}
    \label{fig:InfoDrop CELoss}
    }
    \subfigure[Our PLACE dropout]{
    \begin{minipage}[t]{0.23\linewidth} 
    \centering
    \includegraphics[scale=0.22]{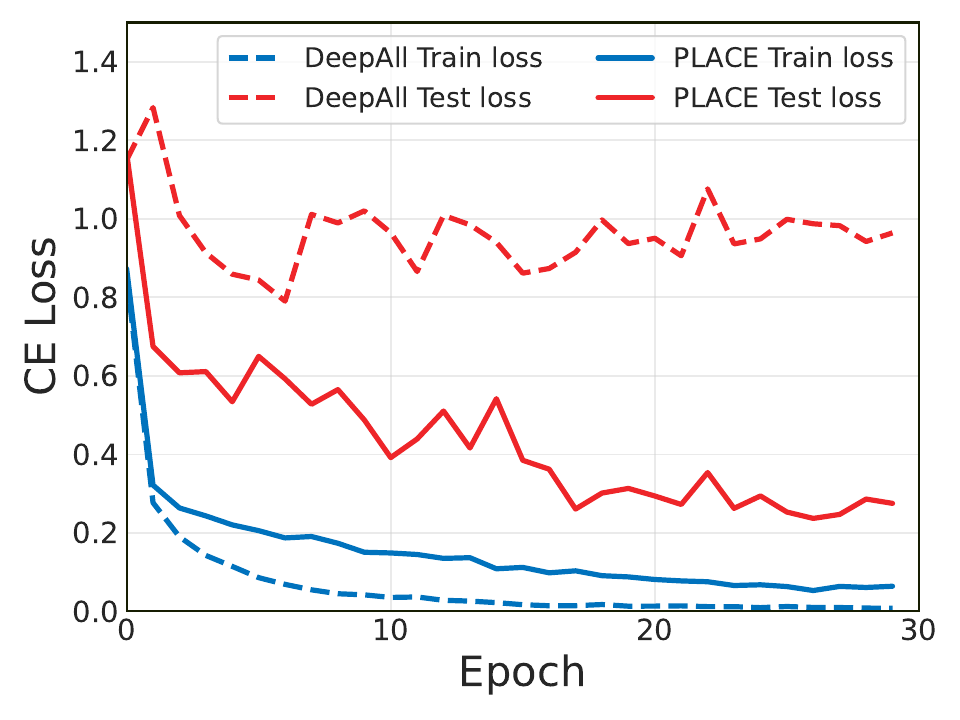}
    \end{minipage}
    \label{fig:PLACE CELoss}
    }
    \centering
    \label{fig:CEloss domain gap}
    \caption{The difference between the training loss and the test loss indicates the overfitting degree. The larger the difference, the more severe the overfitting issue of the model on source domains. We experiment on the PACS dataset with ResNet-$18$ as the backbone architecture. We select the conventional DG method DeepAll \cite{zhou2020deep} and two representative layer-specific dropout methods, \textit{i.e.}, RSC \cite{huang2020self} and InfoDrop \cite{shi2020informative}. Our PLACE dropout can produce regularization effects on multiple layers of the network, which alleviates overfitting effectively with a decreased gap between source train loss and target test loss as shown in column $4$.}
\end{figure}

One of the main challenges in DG is that the model is prone to overfitting the source domains due to the domain gap between source domains and unknown target domain, which greatly impairs the ability of the model to generalize to the target domain. 
One way to identify whether overfitting occurs is to examine if the difference between the training loss and the test loss is increasing \cite{huang2021rda}.
As shown in Fig.~\ref{fig:DeepAll CELoss}, the conventional DG method, \textit{i.e.}, Deepall \cite{zhou2020deep}, overfits the observed source domains soon after training begins due to the large distribution discrepancy between source and target domains.
To alleviate the overfitting, some dropout-based regularization methods \cite{shi2020informative,huang2020self} have shown their promising performance in DG task. 
Without introducing extra parameters, these methods can still achieve remarkable improvements over other DG methods by discarding potentially overfitting-related features during training. 
However, exiting dropout-based DG methods are mainly designed for specific layers, \textit{e.g.}, InfoDrop \cite{shi2020informative} only works in the shallowest layer to reduce the texture-bias of the model, and RSC \cite{huang2020self} only fits in the fully connected layer to reduce the model dependence on the over-dominate features.
As a result, these methods could not produce a sufficiently regularizing effect
on the model, leading to the relatively poor generalization ability to unseen target domain if the model suffers a severe overfitting issue on source domains as shown in \ref{fig:RSC CELoss} and \ref{fig:InfoDrop CELoss}.
Therefore, different from previous layer-specific dropout methods, we design a multiple-layers dropout method for DG to alleviate the overfitting issue adequately.

To be specific, we propose a novel Progressive LAyer-wise and ChannEl-wise (\textbf{PLACE} in short) dropout for domain generalization. 
The PLACE dropout consists of two components, \textit{i.e.}, 1) the layer-wise and channel-wise dropout for perturbing the feature maps in multiple layers, and 2) the progressive scheme to gradually increase the difficulty of training the model.
1) \textit{Layer-wise and channel-wise dropout.} 
Supported by previous dropout-related theoretical works \cite{zeiler2014visualizing,zhang2021can}, we argue that conducting dropout in multiple layers could produce a stronger regularization effect than the single-layer dropout.
However, conducting dropout in multiple layers simultaneously suffers a high risk of losing excessive information, which will most likely hinder model training and slow down the learning speed of the lower layers \cite{park2016analysis}.
To overcome these issues, we design a simple \textit{layer-wise and channel-wise dropout} which randomly selects a layer and then randomly drops its channels at each iteration. 
Our method can effectively resist the overfitting problem by generating diverse data variants in multiple layers during training.
We also theoretically analyze the property of \textit{layer-wise and channel-wise dropout} and prove that it can achieve a small generalization error on target domains. 
2) \textit{Progressive scheme.} 
Considering the risk of overfitting is small at the beginning of training but increases as training progresses \cite{morerio2017curriculum}, 
we introduce a training strategy that enhances the dropout ratio with a progressive scheme, which gradually raises the regularization effect to better tackle the overfitting issue.
Extensive experiments validate the effectiveness of both the proposed dropout and the progressive scheme. 
We also conduct an in-depth analysis of the impact of our method on the cross-domain gap, which demonstrates that PLACE dropout can narrow the gap between source and target domains by reducing the impact of overfitting on source domains.

Our contributions can be summarized as follows: 
\begin{itemize}
  \item We propose a novel regularization method for DG, namely \textit{LAyer-wise and ChannEl-wise dropout}, which can generate diverse variants of data in multiple layers to fight against the overfitting issue of the model on source domains. 
  We also provide theoretical analysis that our method can generate a tight generalization error bound.
  \item We design a simple yet effective \textit{progressive scheme} that gradually boosts the ratio of the layer-wise and channel-wise dropout to continually
  raise the difficulty of training the model, which can further improve the generalization ability of the model.
  \item Without introducing additional network parameters, our method achieves state-of-the-art accuracy on multiple benchmarks and outperforms all established dropout-based methods. 
  We also build a strong baseline for DG with multiple augmentation methods, with which our PLACE dropout can achieve new SOTA performances, $e.g.$, $89.03\%$ on PACS with ResNet-$50$.
\end{itemize}

\section{Related Work}

\subsection{Domain Generalization}
Domain generalization (DG) aims to learn a robust model from multiple distinct but related domains that can generalize well to arbitrary unseen target domains.
Existing DG methods can be roughly divided into four categories, including domain-invariant learning, meta-learning methods, data augmentation and regularization methods.
Several works for domain-invariant learning have been explored, including domain-adversarial learning \cite{fan2021adversarially} and feature disentanglement \cite{chen2021style}.
Another popular way to address the DG problem is meta-learning, which simulates the domain shift by splitting the source domains into meta-train and meta-test domains \cite{dou2019domain,zhang2022more,wei2021metaalign,zhang2022mvdg}.
Data augmentation is also an important technology to empower the model generalization by enriching the diversity of existing training data from the image or feature level.
The image-level augmentation methods mainly generate virtual images via gradient-based adversarial attacks \cite{volpi2018generalizing}, domain-adversarial generation \cite{shu2021open} or learnable augmentation networks \cite{zhou2020deep}.
And the feature-level augmentation methods can diversify image style by mixing or perturbing the feature statistics,
thus generating data in different styles from source domains to boost model generalization \cite{wang2022feature,li2021simple,zhang2023domainadaptor}. 
Recently, some regularization methods have also been
proposed to address the DG task via dropout strategies \cite{shi2020informative,huang2020self}, shape-biased learning \cite{nam2021reducing} and self-supervise methods \cite{carlucci2019domain,wang2020learning}.

Our work is most relevant to the dropout-based regularization methods \cite{shi2020informative,huang2020self}, 
which have shown promising performance for significantly enhancing the model generalization ability with introducing no extra parameters.
Concretely, Shi \textit{et al}. \cite{shi2020informative} find less informative regions contain texture-biased representation, thus proposing to mask them during training to reduce the model's texture bias.
Huang \textit{et al}. \cite{huang2020self} assume that representations with the highest gradients are over-dominant and hinder the model from generalizing, thus designing a self-challenging algorithm to discard them during training.
However, these assumption-based dropouts only apply to specific layers, which is insufficient to regularize the model if there exists a severe overfitting issue on source domains \cite{park2016analysis,srivastava2014dropout}. 
In contrast to all the methods above, we investigate the role of dropout in model training and propose an assumption-free dropout method with layer-wise and channel-wise dropout and a progressive scheme to tackle the DG issue. 
Our method can be applied to multiple layers of the network to generate various data variants during training, which can adequately mitigate the overfitting problem of the model on source domains.

\subsection{Dropout Regularization}
Dropout \cite{srivastava2014dropout} is one of the most widely employed regularization techniques to enhance the generalization capability of deep neural networks. Over the years, dropout has been extended in both channel-wise and space-wise manners. Tompson \textit{et al}. \cite{tompson2015efficient} introduced Spatial Dropout, which drops entire channels in feature maps, while Ghiasi \textit{et al}. \cite{ghiasi2018dropblock} proposed DropBlock, a method that randomly masks contiguous feature regions.
Subsequently, various novel dropout methods have been investigated, broadly categorized into task-auxiliary dropout and structure-information dropout. Task-auxiliary dropout methods aim to incorporate auxiliary tasks to help the model focus on relevant information for the primary task \cite{nagpal2020attribute,schreck2019idropout}, \textit{e.g.}, Nagpal \textit{et al}. \cite{nagpal2020attribute} employ an auxiliary network to drop filters responsible for encoding given sensitive attributes. \textcolor{revisecolor}{On the other hand, structure-information dropout methods leverage feature-level structure information to guide the dropout operations during training for diverse tasks \cite{hou2019weighted,zeng2021correlation,liu2023exploring}}.

However, these methods are primarily designed for supervised and semi-supervised learning settings, where training and test data typically share similar distributions.
Consequently, they may not be well-suited for Domain Generalization (DG) tasks due to their insufficient regularization effect in combating the severe overfitting issues caused by the domain gap between source and target domains.
Moreover, since the target domain is unavailable during training, unsuitable guidance for dropout might potentially damage the generalization ability of the model \cite{zeng2021correlation}. 
To address these challenges, we propose a novel dropout-based framework for DG, which can effectively mitigate the overfitting on source domains, reducing the cross-domain gap, and facilitating the model to generalize effectively to unseen target domains.

\begin{figure*}[tb!]
        \centering
        \includegraphics[width=\linewidth]{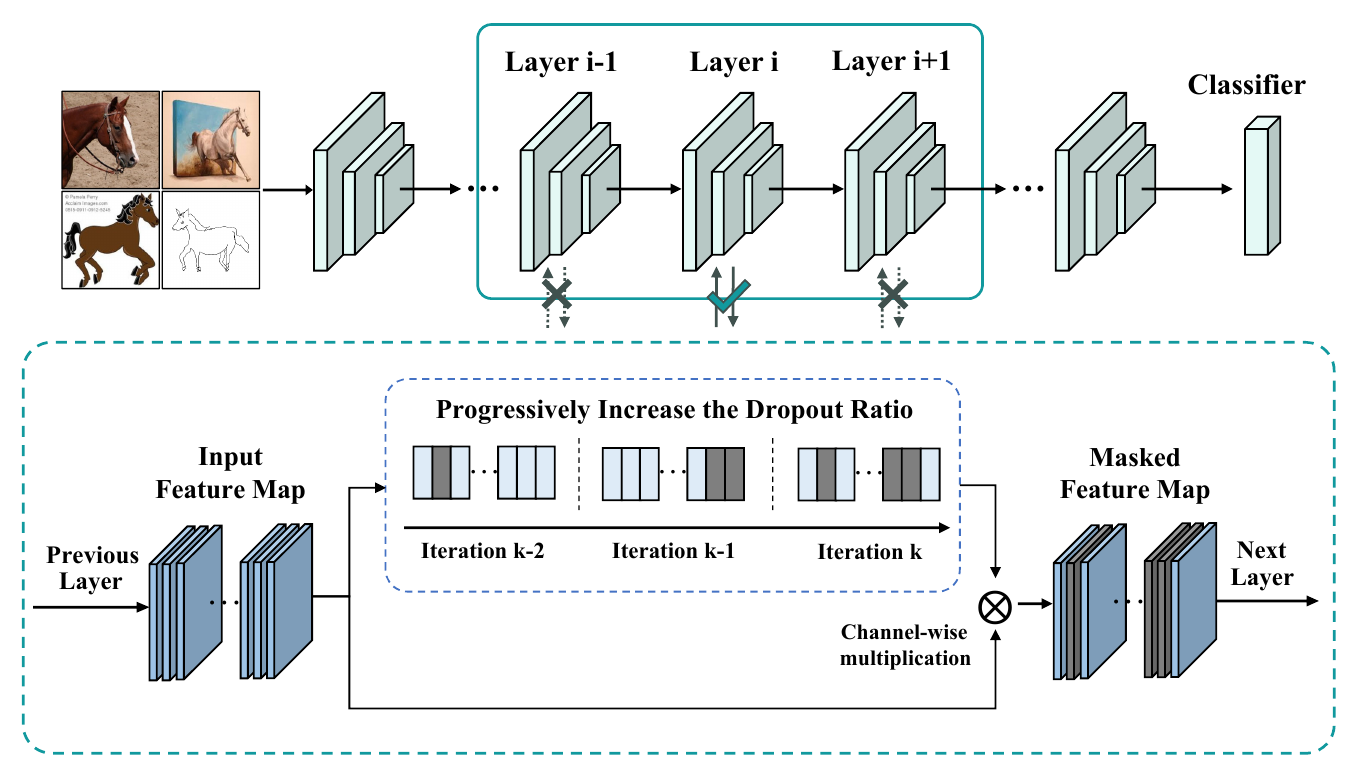}
        \caption{
            Illustration of our PLACE dropout. Our method contains two primary components, including the layer-wise and channel-wise dropout and the progressive scheme. 
            The figure shows the model is trained at the $k$-th iteration and conducted dropout in the $i$-th Layer. Detailed procedure is discussed in Section~\ref{section: training algorithm}.
        }
        \label{figure: Framework}
    \end{figure*}

\section{Proposed Method}
In this section, we introduce the proposed method, Progressive LAyer-wise and ChannEl-wise (PLACE in short) dropout. 
We first present the necessary preliminaries of our work.
Then, we provide the details of the proposed method, including the layer-wise and channel-wise dropout and the progressive scheme.
Finally, we outline the whole training algorithm of our PLACE dropout.

\subsection{Preliminaries}
Given a training set of multiple observed source domains $\mathcal{D}_S = \{D_1, D_2, ..., D_K\}$ with $N_k$ labelled samples $\{(x_i^k, y_i^k)\}_{i=1}^{N_k}$ in the $k$-th domain $S_k$, where $K$ is the number of total source domains, $x_i^k$ and $y_i^k$ denote the samples and labels, respectively.
For simplicity, we use $(\textbf{x}, \textbf{y})$ to replace the sample-label pair $(x_i^k, y_i^k)$ in the following.
The goal of domain generalization is to learn a domain-agnostic model $f(\cdot; \theta)$ on source domains $\mathcal{D}_S$ that can generalize well to any unseen target domain $\mathcal{D}_T$ without extra training.
$\theta$ denotes the network parameters of $f(\cdot; \theta)$.

With the training data of source domains $\mathcal{D}_S$, our method minimizes the standard loss function:  
\begin{equation}
    \mathcal{L}_{ent} = \sum_{\langle \textbf{x},\textbf{y} \rangle \sim \mathcal{D}_s} \ell(f(\textbf{x}; \theta), \textbf{y}).
\end{equation}
where $\ell(\cdot, \cdot)$ is the cross entropy loss function.
However, unlike previous DG methods, PLACE dropout focuses on generating noise by applying dropout to multiple layers for boosting the model robustness and minimizing the above loss function.
Our method consists of two components, \textit{i.e.}, \textit{layer-wise and channel-wise dropout} to generate diverse data variants in multiple network layers,
and \textit{progressive scheme} for increasing dropout ratio to gradually raise the difficulty of training model.
Concretely, at each iteration, PLACE dropout first randomly selects one middle layer, then randomly mutes its channels by a proportion that gradually increases as the training progresses, and finally updates the entire model.
The overview of our method is illustrated in Fig.~\ref{figure: Framework}. 
We present the proposed method in detail in the following parts.

\subsection{Progressive Layer- and Channel-wise Dropout}

\textbf{Channel-wise Dropout.} 
\label{section: channel-wise dropout}
We utilize channel-wise dropout \cite{tompson2015efficient},hich randomly discards entire channels in feature maps, rather than the standard dropout \cite{srivastava2014dropout} or the spatial-wise dropout \cite{ghiasi2018dropblock}.
The reason behind this choice lies in the high similarity between adjacent units within feature maps. Standard dropout, which drops individual neurons, does not exhibit a marked effect in convolution layers \cite{park2016analysis}. 
On the other hand, spatial-wise dropout, which masks contiguous regions, faces challenges in its regularization effect, as it is easily influenced by the location and size of the dropped regions, \textit{i.e.}, over-dropping if the informative regions (\textit{e.g.}, foreground regions) are masked or under-dropping if the irrelevant regions (\textit{e.g.}, background regions) are discarded \cite{ghiasi2018dropblock,zeng2021correlation}. 
In contrast, channel-wise dropout operates at the level of individual channels, which are basic units of feature maps corresponding to specific patterns in the input image. 
As a result, channel-wise dropout has been demonstrated to effectively ensure the dropout's impact in convolution layers \cite{tompson2015efficient,hou2019weighted}. 
Besides, dropping channels in feature maps reduces co-adaptations among different channels \cite{srivastava2014dropout,tompson2015efficient}, which can effectively mitigate the overfitting issue on source domains and implicitly encourage the model to learn comprehensive feature patterns.

\vspace{0.1cm}
\noindent\textbf{Layer-wise and Channel-wise Dropout.} 
\label{layer-wise dropout}
Different from most previous dropout-based DG methods by conducting the dropout on specific layers \cite{shi2020informative,huang2020self}, we propose
layer-wise dropout that randomly selects a middle layer of the network and then randomly discards its channels at each iteration.
We first analyze the effect of random channel-wise dropout in different layers on the inter-domain discrepancy between source and target domains, which potentially reflects the overfitting degree of the model to the observed source domains. The discrepancy is calculated as \cite{wang2022feature}:
\begin{equation}
    d^{i} = \frac{1}{K} \sum_{s=1}^{K} || {\rm GAP}(\overline{\textbf{z}}_s^i) - {\rm GAP}(\overline{\textbf{z}}_t^i)  ||_2,
    \label{eq:inter-domain instance}
\end{equation}
where $\overline{\textbf{z}}_s^i$ is the averaged feature maps of all samples from the $i$-th layer in the $s$-th source domain, and $\overline{\textbf{z}}_t^i$ is the mean feature maps from the $i$-th layer in the target domain.
$K$ is the number of source domains. ${\rm GAP}(\cdot)$ is the global average pooling operation. 
We calculate the domain discrepancy $d_{drop_{j}}^{i}$ according to features from the $i$-th layer during inserting dropout into the $j$-th layer, and $d_{bal}^{i}$ for the baseline model. Then we calculate $d_{drop_{j}}^{i} - d_{bal}^{i}$ to investigate the effectiveness of dropout, where the ``negative" value indicates that the dropout can reduce the domain difference.

\begin{figure}[htbp!]
  \centering
  \subfigure[PACS]{
  \begin{minipage}[t]{0.45\linewidth}
  \centering
  \includegraphics[scale=0.36]{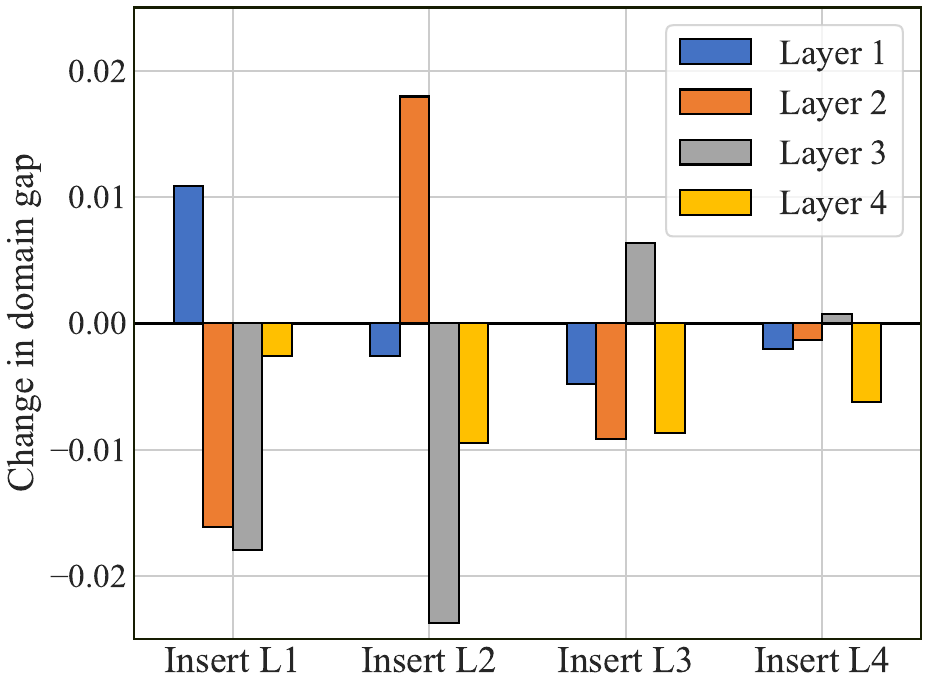} 
  \end{minipage}
  \label{fig:PACS domain gap}
  }
  \subfigure[OfficeHome]{
  \begin{minipage}[t]{0.45\linewidth}
  \centering
  \includegraphics[scale=0.36]{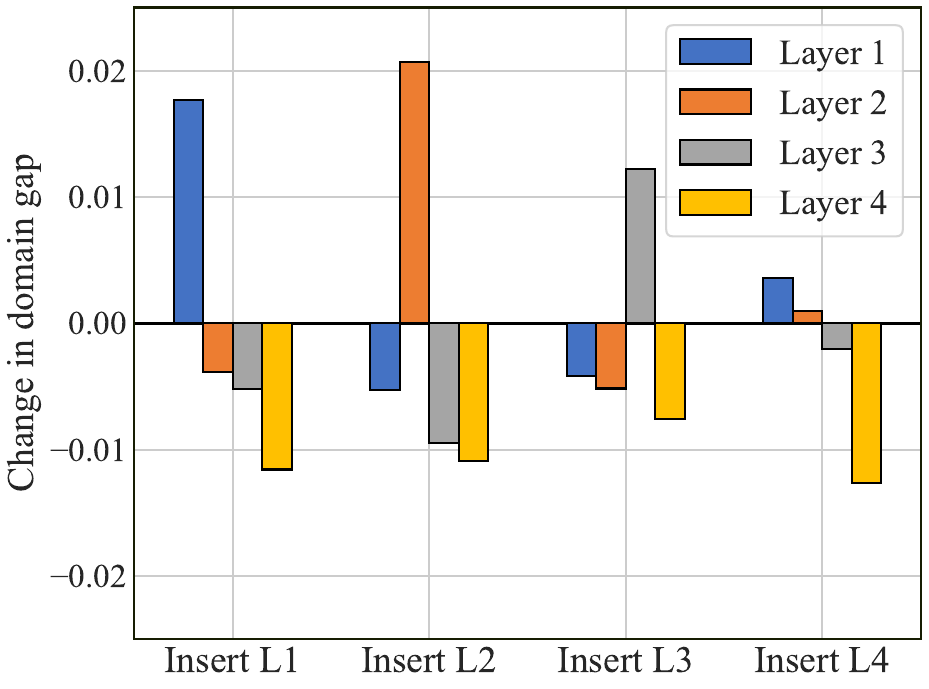}
  \end{minipage}
  \label{fig:OfficeHome domain gap}
  }
  \centering
  \vspace{-0.2cm}
  \caption{The effectiveness of dropout in each layer of ResNet-$18$ when dropout is inserted in different layers. 
  The horizontal axis denotes the position of inserting channel-wise dropout (\textit{i.e.}, $L_i$ is the $i$-th layer), and
  the vertical axis is the magnitude of change in domain gap from the $i$-th layer, which is calculated by $d_{drop}^{i} - d_{bal}^{i}$.}
  \label{fig:layer domain gap}
  \vspace{-0.2cm}
\end{figure}

The results are presented in Fig.~\ref{fig:layer domain gap}. Specifically, Fig.~\ref{fig:PACS domain gap} presents the results on PACS when considering Art as the target domain and using the other three domains to train the model, and Fig.~\ref{fig:OfficeHome domain gap} presents the results on OfficeHome when taking RealWorld as the target domain.
From Fig.~\ref{fig:layer domain gap}, we observe that regardless of the layer where dropout is inserted, it consistently leads to a lower domain gap between the source and target domains at the $4$-th layer, which is nearest to the prediction layer and its output can directly influence the prediction performance of the model.
Furthermore, the results suggest that placing dropout in different layers has different effects on the domain gap between the source and target domains in each layer, \textit{i.e.}, leading to a larger domain gap in the inserted layer but a smaller domain gap in the adjacent layers.
This observation could be interpreted as that dropout introduces noise to the feature maps of the inserted layer, compelling the adjacent layers to learn general and robust features to resist the noise.
In conclusion, \textit{the feature representations in different layers exhibit significant diversity, leading to various effects on model training when dropout is placed in different layers.}

However, conducting dropout in multiple layers simultaneously could lead to conflicts, \textit{e.g.}, if dropout is set in the $2$-nd and $3$-rd layers, dropout in the $2$-nd layer will reduce the domain discrepancy on $3$-rd layer, but setting dropout in $3$-rd layer will increase the domain discrepancy itself.
Moreover, the simultaneous usage of dropout in multiple layers could result in the absence of excessive information during training and hinder the model convergence \cite{park2016analysis}.
To address these issues, we propose a novel approach called \textit{layer-wise and channel-wise dropout}. Instead of applying dropout to all layers at once, our method randomly selects a middle layer of the network and performs dropout on its feature maps at each iteration. 
This approach introduces noise at multiple scales to the model, generating various data variants to enhance model robustness.
Furthermore, our proposed PLACE dropout can effectively integrate the effects of dropout in different layers while avoiding conflicts in the inter-domain discrepancy across layers, thus producing a strong regularization effect to mitigate the overfitting issue on source domains.

\vspace{0.1cm}

\noindent\textbf{Progressive Scheme.} 
We design a time-adaptive regularization method, namely \textit{progressive scheme}, to mitigate the overfitting issue adequately while ensuring the stability of model convergence. 
It is motivated by the intuition that the risk of overfitting on source domains is low at the beginning of training, and increases as the training proceeds \cite{morerio2017curriculum}. 
To address this issue, we propose to increase the dropout ratio progressively, which can gradually enhance the regularization effect of dropout \cite{wager2013dropout} and help the model better tackle the overfitting problem on source domains.
Given the current iteration $t$, 
we compute the dropout ratio $p$ and the number of dropped channels $\gamma$:
\begin{equation}
    \gamma= C \times p = C \times P_{max} \times G (t), 
    \label{eq:Progressive formula}
\end{equation}
\begin{equation}
    G(t)=\frac{2}{\pi} \arctan(\frac{t}{V}),
    \label{eq:G formula}
\end{equation} 
where $P_{max}$ denotes the maximum of channel-wise dropout ratio, $C$ represents the number of channels, and $V$ is a hyper-parameter to control the increasing speed of $p$, respectively.
$G(\cdot)$ could be an appropriate monotonically increasing and strongly-convex function with $t$ as input and a value from $0$ and $1$ as output, \textit{i.e.}, the growth rate is relatively high at the start of training but gradually decreases as the training processes.
Such function is also utilized in curriculum learning and related progressive settings \cite{morerio2017curriculum,bengio2009curriculum}.
We empirically adopt the arctangent function (\textit{i.e.}, $\arctan(\cdot)$) for $G(\cdot)$, which is formulated as Eq. (\ref{eq:G formula}), in the following experiments.


\begin{algorithm}[tb!]
    \caption{PLACE dropout Algorithm}
    \label{algorithm:PLACE}
    \KwIn {Batch size $N$, learning rate $\eta$, source data $\{(x_{k}, y_{k})\}$, candidate layer set $\{l_{1}, \dots, l_{n}\}$, updated network $f_{\theta_{u}}$.}
    \KwOut {Trained network $f_{\theta_{u}^{*}}$.}
    \For{sampled mini-batch $\{(x_{k}, y_{k})\}^{N}_{k=1}$}{
        Sample $l$ from $\{l_{1}, \dots, l_{n}\}$; \hspace{2em} \textcolor{commentcolor}{{\small // Select one middle layer from the network.}} \\
        $\textbf{F}^{l} \in \mathbb{R} ^{C \times H \times W}  \leftarrow$ Computed by forward propagation; \\
        $P$ and $\gamma \leftarrow$ Computed by Eq. (\ref{eq:Progressive formula}); \hspace{2em} \textcolor{commentcolor}{{\small // Compute the proportion of dropout.}} \\
        \For{each sample $(x_{i}, y_{i})$}{
            Randomly sample $\gamma$ channels $\{r_1, r_2, \cdots, r_{\gamma}\}$ from $F_{i}^{l}$; \\
            $\textbf{M}^{l}_{i} \leftarrow$ Computed by Eq. (\ref{eq:channel mask}); \hspace{2em} \textcolor{commentcolor}{{\small // Compute the channel-wise mask matrix.}} \\
            $\widehat{\textbf{F}}^{l} \leftarrow$ Computed by Eq. (\ref{eq:refined feature});  \hspace{2em} \textcolor{commentcolor}{{\small //  Compute the masked feature map.}} \\
        }
        Update $\textbf{F}^{l} \leftarrow \widehat{\textbf{F}}^{l}$; \hspace{2em} \textcolor{commentcolor}{{\small // Update current feature map to the masked version.}} \\
        Compute the output of the succeeding layers; \hspace{2em} \textcolor{commentcolor}{{\small // Forward propagation.}} \\
        Compute $L \leftarrow -\frac{1}{N} \sum_{i=1}^{N} \sum_{k=1}^{K} y_{i k} log(p_{i k})$; \hspace{2em} \textcolor{commentcolor}{{\small // Compute the prediction loss.}} \\
        Update $\theta_{u} \leftarrow \theta_{u} - \eta \nabla_{\theta_{u}} L$; \hspace{2em} \textcolor{commentcolor}{{\small // Update the parameters of the network.}} \\
    }
  \end{algorithm}

\subsection{Training Algorithm}
\label{section: training algorithm}
As introduced before, at each iteration, we first randomly select an index $l$ of a middle layer from the network.
Then we obtain the output of the $l$-th layer by forward propagation, which is denoted as $\textbf{F}^{l}\in \mathbb{R} ^{C \times H \times W}$, and generate an all-one mask matrix $\textbf{M}^{l} \in \mathbb{R} ^{C \times H \times W}$ with the same size as $\textbf{F}^{l}$. $C$ is the number of channels, $H$ and $W$ denote the dimension of height and width, respectively. 
Next, we calculate the dropout proportion $p$ according to current iteration $t$ by Eq. (\ref{eq:Progressive formula}) and randomly sample $\gamma$ distinct channel indexes $\{r_1, r_2, \cdots, r_{\gamma}\}$ from the $C$ channels of $\textbf{M}^{l}$.
The element in $\textbf{M}^{l}$ is set to $0$ if its corresponding index is one of the selected channel indexes, and set to 1 otherwise:
    \begin{equation}
    \textbf{M}^{l}_{\textit{i, j, k}} = 
    \begin{cases}
    0,& \textit{i} \in \{r_1, r_2, \cdots, r_{\gamma}\} \\
    1,& \text{otherwise}
    \end{cases}.
    \label{eq:channel mask}
    \end{equation}

We then obtain the masked output features map $\widehat{\textbf{F}}^{l}$ as Eq. (\ref{eq:refined feature}) by computing element-wise multiplication (denoted as $\odot$) of the output $\textbf{F}^{l}$ and the mask matrix $\textbf{M}^{l}$:
\begin{equation}
    \widehat{\textbf{F}}^{l}=\textbf{F}^{l} \odot \textbf{M}^{l}.
    \label{eq:refined feature}
\end{equation}
Finally, $\widehat{\textbf{F}}^{l}$ is forward propagated to the following part of the network to compute the prediction loss and update the parameters of the entire network.

The overall training procedure of PLACE dropout is summarized in Algorithm \ref{algorithm:PLACE}. 
In the training stage, 
our method can generate diverse data variants in multiple layers for tackling the overfitting issue on source domains.
During the inference process, our PLACE dropout is closed as the conventional dropout \cite{srivastava2014dropout}.
It's worth noting that the PLACE dropout only comprises a few simple operations, such as random selection and channel-level product, \textit{i.e.}, our method introduces no additional parameters and only incurs negligible computational cost during training.
It also does not consume any extra computation time in the inference stage.

\section{Theoretical Analysis}
In this section, we provide the theoretical analysis for layer-wise and channel-wise dropout and prove that our method can effectively reduce the generalization error bound of the model.
We first analyze the domain generalization error bound for the dropout-based model, which can help us analyze the relationship between multi-layers dropout and model generalization.
Then we demonstrate that under mild assumptions, 1) \textit{layer-wise and channel-wise dropout} can reduce the model sensitivity to the perturbation caused by dropout.
2) \textit{layer-wise and channel-wise dropout} can generate more diverse augmented data than single-layer dropout; 
The results indicate that \textit{layer-wise and channel-wise dropout} can lower the upper bound of generalization error and help the model generalize well to arbitrary unseen target domains.

\vspace{0.1cm}

\noindent\textbf{Notations.} 
Given the sample-label pair (\textbf{x}, \textbf{y}), we use $\textbf{z}$ to denote the feature representation of (\textbf{x}, \textbf{y}) learned by the model.
We define the task component of the model as $h: \mathcal{Z} \rightarrow \mathcal{Y}$ such that $h \in \mathcal{H}$, where $\mathcal{H}$ is a set of candidate hypothesis.
$h$ takes $\textbf{z}$ as input and outputs the corresponding predict label.
For simplicity, we take $S$ to represent the set of feature-label pairs $\{(\textbf{z}, \textbf{y})\}$.
$\widetilde{\textbf{z}}$ is the perturbed version of $\textbf{z}$ generated by dropout.
$\widetilde{S}$ is the set containing all possible perturbed versions $\widetilde{\textbf{z}}$ of representations $\textbf{z}$ in $S$.
Given a hypothesis $h$, the empirical risk $R[h]$ on domain $\mathcal{D}$ is defined:
\begin{equation}
    R_D[h] = \mathbb{E}_{\langle \textbf{z}, \textbf{y} \rangle \sim \mathcal{D}} \ell[h(\textbf{z}), \textbf{y}],
    \label{eq:risk definition}
\end{equation}
where the loss: $\ell: \mathcal{Y} \times \mathcal{Y} \rightarrow R_+$ quantifies the difference between $h(\textbf{z})$ and $\textbf{y}$ for a data pair $(\textbf{z}, \textbf{y})$.

\vspace{0.1cm}

We first introduce the generalization error bound of the dropout-based model under the DG set. 
Assuming that samples in the unknown target domain $T$ could be regarded as perturbed versions of samples in source domains $S$, the empirical risk on the target domain is upper-bounded \cite{huang2020self} by:

\vspace{0.1cm}

\noindent\textbf{Lemma 1 \cite{huang2020self}.} \textit{
Let $L(\textbf{z}_S, h)$ be the loss of a given hypothesis $h$ on the source domains $S$, defined as:
\begin{equation}
    L(\textbf{z}_S, h) = \mathbb{E}_{\langle \textbf{z}, \textbf{y} \rangle \sim \mathcal{S}} \ell[h(z), y].
    \label{eq:loss}
\end{equation}
$\xi(h)$ is a function of $h$ that can reflect the sensitivity of the model to the perturbation of dropout:
\begin{equation}
    \xi(h) = \sup_{\widetilde{\textbf{z}}_S \in \widetilde{S}} |L(\textbf{z}_S, h) -  L(\widetilde{\textbf{z}}_S, h)|,
\label{eq:loss difference}
\end{equation}
where $\widetilde{\textbf{z}}$ is the perturbed version of $\textbf{z}$ by dropout. Let $N$ be the dataset size.
Then for any $h \in \mathcal{H}$ and $\delta \in (0, 1)$, the following inequality holds with the probability at least $1- \delta$:
\begin{equation}
    R_T[h] \leq R_S[h] + \xi(h) \sqrt{\frac{\ln|\mathcal{H}| + \ln(2/\delta)}{2N}}.
    \label{eq:generalization error}
\end{equation}}

The bound at the right side of Eq.~\ref{eq:generalization error} contains two terms: (1) the first is the empirical risk over all source domains; (2) the second indicates that the generalization bound of the model is positively related to the sensitivity $\xi(h)$ of the model to dropout, and negatively related to the dataset size $N$ of source domains. 
We then analyze the effectiveness of layer-wise and channel-wise dropout by presenting that it reduces the sensitivity $\xi(h)$ while increasing the dataset size $N$.

\vspace{0.1cm}

\noindent\textcolor{revisecolor}{\textbf{Proposition 1.} \textit{
    Let $\widehat{\xi}_t(h)$ be the estimated value of $\xi(h)$ at iteration $t$, which is defined as: $\widehat{\xi}_t(h) = |L(\textbf{z}_t, h_t) -  L(\widetilde{\textbf{z}}_t, h_t)|$. 
    Given a sufficiently small learning rate $\eta$, if discarding structural features will increase the empirical loss at the current iteration $t$, 
    it holds that the layer-wise and channel-wise dropout can continually decrease $\widehat{\xi}_t(h)$ and lead it to be a small number at the end of training.
}}
\vspace{0.1cm}

\noindent\textcolor{revisecolor}{\textit{\textbf{Proof.}} See in the Appendix.}

\vspace{0.1cm}
\noindent\textcolor{revisecolor}{\textbf{Proposition 2.} \textit{
Dropout in different layers perform as different data augmentations in the input space, thus the layer-wise dropout can generate more diverse augmented data than the single-layer dropout, \textit{i.e.}, increasing the dataset size $N$.
}}
\vspace{0.1cm}

\noindent\textcolor{revisecolor}{\textit{\textbf{Proof.}} See in the Appendix.}

\vspace{0.1cm}

According to Propositions $1$ and $2$, we can conclude that \textit{the layer-wise and channel-wise dropout can enrich the diversity of training data and reduce the sensitivity of model to the perturbation by dropout, thus generating a tight generalization error bound. }
The theoretical results justify our motivation and formally verify the efficacy of our method in terms of improving the generalization error bound. 
In the following, we will experimentally demonstrate the superiority of our method.

\section{Experiments and Results}
\subsection{Datasets and Settings}
\textbf{Datasets.} We evaluate our method on three popularly-used DG benchmark datasets as follows: 
\begin{itemize}
    \item \textbf{PACS} \cite{li2017deeper} consists of images from $4$ domains with a large discrepancy in image styles: Photo, Art Painting, Cartoon, and Sketch, including $7$ object categories and $9,991$ images total. We adopt the official split provided by \cite{li2017deeper} for training and test; 
    \item \textbf{VLCS} \cite{torralba2011unbiased} comprises of 5 categories, which are selected from $4$ photo datasets, \textit{i.e.}, VOC $2007$ (Pascal), LabelMe, Caltech and Sun datasets. We utilize the same experimental setup as \cite{carlucci2019domain} and divide the dataset into the training and test sets based on $7:3$; 
    \item \textbf{Office-Home} \cite{venkateswara2017deep} is an object recognition benchmark that contains around $15,500$ images of $65$ categories from $4$ domains: Artistic, Clipart, Product and Real-World. Following \cite{carlucci2019domain}, we randomly split each domain into $90\%$ for training and $10\%$ for test.
\end{itemize}

\textbf{Experimental Settings.}
We apply the leave-one-domain-out protocol for all benchmarks, \textit{i.e.}, we train the model on source domains and test the model on the remaining domain.
Due to the progressive scheme, we select the model of the last epoch as the final model and report the top-$1$ classification accuracy.
All the reported results are the averaged value over five runs. 

\subsection{Implementation Details}
\label{Implementation Details}
For the PACS dataset, we use the ImageNet pre-trained ResNet-18 and ResNet-50 as backbones following \cite{he2016deep} and adopt the same hyper-parameters as \cite{carlucci2019domain}. 
For the VLCS dataset, We employ the experimental protocol as mentioned in \cite{zhang2021deep} and use ResNet-18 as the backbone.
For the OfficeHome dataset, we follow the same experimental setup as \cite{d2018domain}.
We train the network using SGD with a momentum of $0.9$ and weight decay of $5 \times 10^{-4}$ for a total of $30$ epochs.
The initial learning rate of the network is $4 \times 10^{-3}$ and decayed by $0.1$ at $80\%$ of the total epochs. 

Our method consists of two primary components, \textit{i.e}., the progressive scheme and the layer-wise and channel-wise dropout. 
For the progressive scheme, the dropout proportion is related to two hyperparameters, \textit{i.e}, the maximum dropout ratio $P_{max}$, and the progressive rate $V$. 
The maximum dropout ratio $P_{max}$ is set as $0.33$ for PACS and VLCS, and $0.25$ for OfficeHome, respectively. 
The progressive rate $V$ is set to $4$ in all datasets.
For the layer-wise and channel-wise dropout, we select the $1$st, $2$nd, and $3$rd residual layers as the candidate set for both ResNet-18 and ResNet-50, which is determined experimentally in Tab.~\ref{table: Layers}. 
At each iteration, we randomly select a residual layer from the candidate set and perform the progressive channel-wise dropout on its feature maps. 
Practically, we design a strong baseline model, denoted as ${\rm DeepAll}^{++}$, for domain generalization with image-level and feature-level augmentation methods, \textit{i.e}., we perform random augmentation \cite{cubuk2020randaugment} on the images before training, and then randomly swap the style statistics of features in the $3$-rd layer for any two samples during training \cite{borlino2021rethinking}, which is motivated by AdaIN \cite{huang2017arbitrary}. 
The strength of random augmentation is controlled by both the number of transformations and the magnitude of distortion, which are set to $8$ and $4$, respectively.

\subsection{Comparison with State-of-the-Art Methods}
We compare our method with other recent state-of-the-art (SOTA) domain generalization methods on three public standard benchmark datasets, \textit{i.e.}, PACS, VLCS, and OfficeHome. 

\renewcommand{\thefootnote}{\fnsymbol{footnote}}

\begin{table}[tb!]
    \begin{center}
    \caption{Comparison of performance (\%) among different methods using ResNet-18 on PACS \cite{li2017deeper}. 
    The best and second-best are \textbf{bolded} and \underline{underlined}, respectively.
    }
    \label{table: PACS}
    \renewcommand\arraystretch{1.1}
    \scalebox{1.0}{
    \begin{tabular}{l | c | cccc |c}
        \toprule
        Method & Venue & Art & Cartoon & Sketch & Photo & Avg.\\
        \midrule
        C-Drop \cite{morerio2017curriculum}\footnotemark[2] & ICCV 2017 & 79.64 & 76.49 & 72.37 & 95.93 & 81.11 \\
        WCD \cite{hou2019weighted}\footnotemark[2] & AAAI 2019 & 81.56 & 78.24 & 75.53 & 94.99 & 82.58 \\
        RSC \cite{huang2020self}\footnotemark[2] & ECCV 2020 & 82.03 & 77.39 & 75.64 & 95.63 & 82.67 \\
        I-Drop \cite{shi2020informative}\footnotemark[2] & ICML 2020 & 80.27 & 76.54 & 76.38 & 96.11 & 82.33 \\
        \midrule
        DSON \cite{seo2020learning} & ECCV 2020 & 84.67 & 77.65 & 82.23 & 95.87 & 85.11\\
        SFA-A \cite{li2021simple}& ICCV 2021 & 81.20 & 77.80 & 73.70 & 93.90 & 81.70\\
        MixStyle \cite{zhou2020domain} & ICLR 2021 & 84.10 & 78.80 & 75.90 & 96.10 & 83.70 \\
        pAdaIN \cite{nuriel2021permuted} & CVPR 2021 & 81.74 & 76.91 & 75.13 & 96.29 & 82.51 \\
        SagNet \cite{nam2021reducing} & CVPR 2021 & 83.58 & 77.66 & 76.30 &  95.47 & 83.25 \\
        FACT \cite{xu2021fourier} & CVPR 2021 & \underline{85.37} & 78.38 & 79.15 & 95.15 & 84.51 \\
        StableNet \cite{zhang2021deep} & CVPR 2021 & 81.74 & \underline{79.91} & 80.50 & \underline{96.53} & 84.69 \\
        EFDMix \cite{zhang2022exact} & CVPR 2022 & 83.90 & 79.40 & 75.00 & \textbf{96.80} & 83.90 \\
        StyleNeophile \cite{kang2022style} & CVPR 2022 & 84.41 & 79.25 & 83.27 & 94.93 & 85.47 \\
        IRMCon-IPW \cite{qi2022class} & ECCV 2022 & 81.10 & 77.30 & 76.60 & 95.40 & 82.60 \\
        I$^{2}$-ADR \cite{meng2022attention} & ECCV 2022 & 82.90 & \textbf{80.80} & \underline{83.50} & 95.00 & \underline{85.60} \\
        \midrule
        DeepAll \cite{zhou2020deep} {\scriptsize (\textit{our imple.})} & AAAI 2020 & 80.19 & 77.19 & 73.48 & 95.71 & 81.64 \\
        \textbf{+ PLACE dropout} & \textbf{Ours} & 82.60 & 78.33 & 81.47 & 95.65 & 84.51\\
        \midrule
        DeepAll$^{++}$ & \textbf{Ours} & 83.64 & 78.28 & 81.41 & 96.34 & 84.92 \\
        \textbf{+ PLACE dropout} & \textbf{Ours} & \textbf{85.40} & 79.69 & \textbf{83.97} & 96.23 & \textbf{86.32} \\
        \bottomrule
    \end{tabular}}
    \end{center}
    \end{table}

\textbf{Results on PACS.} 
We compare our PLACE dropout with SOTA methods of both dropout and domain generalization on the PACS dataset. Tab.~\ref{table: PACS} and Tab.~\ref{table: PACS on ResNet-50} present the results using ResNet-$18$ and ResNet-$50$ as backbones, respectively.
In Tab.~\ref{table: PACS}, PLACE dropout significantly outperforms the SOTA dropout-based method (RSC \cite{huang2020self}) by $1.84\%$ ($84.51\%$ vs. $82.67\%$), which develops a self-challenging algorithm by discarding over-dominant features with large gradients to encourage the model to rely more on the remaining features. 
The superior performance of PLACE dropout compared to previous dropout-based methods highlights the importance of utilizing dropout in multiple layers.
Furthermore, when compared to SOTA DG methods, PLACE dropout achieves competitive performance and outperforms the baseline by a significant margin of $2.87\%$ ($84.51\%$ vs. $81.64\%$). 
While integrated with our proposed DeepAll$^{++}$, which incorporates existing image-level and feature-level augmentation methods, PLACE dropout further improves the model generalization and outperforms the SOTA DG method I$^{2}$-ADR \cite{meng2022attention} by $0.72\%$ ($86.32\%$ vs. $85.60\%$).
It is noteworthy that our method achieves excellent performance with introducting no extra parameters and little training time, proving its efficiency and superiority.
Tab.~\ref{table: PACS on ResNet-50} presents the results on PACS using ResNet-$50$ as the backbone. As observed, PLACE dropout continues to achieve substantial improvements over DeepAll by $3.15\%$ ($87.83\%$ vs. $84.68\%$) and DeepAll$^{++}$ by $1.93\%$ ($89.03\%$ vs. $87.10\%$). 
\textcolor{revisecolor}{
    We notice that on DeepAll$^{++}$ in Tab.~\ref{table: PACS on ResNet-50}, PLACE dropout slightly degrades the performance on Cartoon, which could be attributed to the excessive perturbations by the combination of PLACE dropout and the two augmentation methods employed. 
    Considering that different intensities of perturbations could lead to variations in the model performance across different domains (as evidenced by Tab.~\ref{table: Pmax PACS}), the performance of our method on cartoon could be improved by fine-tuning perturbation strength accordingly (\textit{i.e.}, decreasing the dropout rate).
}
Moreover, our approach surpasses the second-best method PCL \cite{yao2022pcl} by $0.33\%$ ($89.03\%$ vs. $88.70\%$), demonstrating the stability and effectiveness of PLACE dropout even when incorporated into a larger network like ResNet-$50$.
\textcolor{revisecolor}{
    Noting that certain methods exhibit slightly better performance than ours on specific domains, , likely due to the utilization of specialized structures or assumptions tailored for those particular domains. 
    In contrast, our method is an assumption-free technique, which is orthogonal to existing methods. The results in Tab.~\ref{table: SOTA + PLACE} further prove that combining existing methods with PLACE dropout can lead to enhanced generalization performance.
}

\footnotetext[2]{The dropout-based regularization methods.}

\begin{table}[tb!]
\begin{center}
\caption{Comparison of performance (\%) among different methods using ResNet-50 on PACS \cite{li2017deeper}.
The best and second-best are \textbf{bolded} and \underline{underlined}, respectively.
}
\label{table: PACS on ResNet-50}
\renewcommand\arraystretch{1.1}
\scalebox{1.0}{
\begin{tabular}{l|c|cccc|c}
    \toprule
    Method & Venue & Art & Cartoon & Sketch & Photo & Avg.\\
    \midrule
    RSC \cite{huang2020self} & ECCV 2020 & 84.13 & 79.83 & 82.32 & 95.35 & 85.41 \\
    EISNet \cite{wang2020learning} & ECCV 2020 & 86.64 & 81.53 & 78.07  & 97.11 & 85.84 \\
    DSON \cite{seo2020learning} & ECCV 2020 & 87.04 & 80.62 & 82.90 & 95.99 & 86.64 \\
    MDGHybrid \cite{mahajan2021domain} & ICML 2021 & 86.74 & 82.32 & 82.66 & \textbf{98.36} & 87.52 \\
    FACT \cite{xu2021fourier} & CVPR 2021 & 89.63 & 81.77 & 84.46 & 96.75 & 88.15 \\
    PCL \cite{yao2022pcl} & CVPR 2022 & \underline{90.20} & \textbf{83.90} & 82.60 & \underline{98.10} & \underline{88.70} \\
    EFDMix \cite{zhang2022exact} & CVPR 2022 & \textbf{90.60} & 82.50 & 76.40 & \underline{98.10}  & 86.90 \\
    I$^{2}$-ADR \cite{meng2022attention} & ECCV 2022 & 88.50 & 83.20  & \underline{85.80} & 95.20 & 88.20 \\
    \midrule
    DeepAll \cite{zhou2020deep} {\scriptsize (\textit{our imple.})} & AAAI 2020 & 85.55 & 79.82 & 76.69 & 96.66 & 84.68 \\
    \textbf{+ PLACE dropout} & \textbf{Ours} & 87.55 & 83.11 & 83.48 & 97.19 & 87.83 \\
    \midrule
    DeepAll$^{++}$ & \textbf{Ours} & 85.11 & \underline{83.23} & 83.58 & 96.47 & 87.10 \\
    \textbf{+ PLACE dropout} & \textbf{Ours} & 89.03 & 83.04 & \textbf{86.82} & 97.22 & \textbf{89.03} \\
    \bottomrule
\end{tabular}}
\end{center}
\end{table}

\textbf{Results on VLCS.} As shown in Tab.~\ref{table: VLCS}, our PLACE dropout outperforms the baseline DeepAll by a significant margin of $4.96\%$ ($77.44\%$ vs. $72.48\%$) and achieves competitive performance with the SOTA method StableNet \cite{zhang2021deep} while introducing no additional training parameters.
Furthermore, based on the DeepAll$^{++}$, our PLACE dropout surpasses the SOTA method StableNet \cite{zhang2021deep} by a margin of $0.11\%$ ($77.76\%$ vs. $77.65\%$) on average, demonstrating the effectiveness of our method.
Specifically, our method outperforms StableNet on the Caltech and Pascal domains with considerable improvements of $1.56\%$ ($98.23\%$ vs. $96.67\%$) and $2.71\%$ ($76.30\%$ vs. $73.59\%$), respectively, but slightly underperforms on the other two domains.
The results could be attributed to the different strategies employed by our method and StableNet. Our method randomly drops task-relevant features to mitigate the overfitting issue, while StableNet discards the task-irrelevant features for stable learning, causing their different effects on the generalization ability of the model.

\begin{table}[tb!]
\begin{center}
\caption{Comparison of performance (\%) among different methods using ResNet-18 on VLCS \cite{torralba2011unbiased}.
The best and second-best are \textbf{bolded} and \underline{underlined}, respectively.
}
\label{table: VLCS}
\scalebox{1.0}{
\renewcommand\arraystretch{1.1}
\begin{tabular}
{l | c | c c c c | c}
\hline
Method & Venue & Caltech & LableMe & Pascal & Sun & Avg.\\
\toprule
    JiGen \cite{carlucci2019domain} & CVPR 2019 & 96.17 & 62.06 & 70.93 & 71.40 & 75.14\\
    MMLD \cite{matsuura2020domain} & AAAI 2020 & 97.01 & 62.20 & 73.01 & \underline{72.49} & 76.18\\
    RSC \cite{huang2020self} & ECCV 2020 & 95.83 & 63.74 & 71.86 & 72.12 & 75.89 \\
    StableNet \cite{zhang2021deep} & CVPR 2021 & 96.67 & \textbf{65.36} & 73.59 & \textbf{74.97} & \underline{77.65}\\
    \midrule
    DeepAll \cite{zhou2020deep} {\scriptsize (\textit{our imple.})}  & AAAI 2020 & 91.86 & 61.81 & 67.48 & 68.77 & 72.48\\
    \textbf{+ PLACE dropout} & \textbf{Ours} & 98.14 & 64.12 & \underline{75.45} & 72.06 & 77.44 \\
    \midrule
    DeepAll$^{++}$ & \textbf{Ours} & \textbf{98.27} & 62.02 & 74.09 & 71.29 & 76.42 \\
    \textbf{+ PLACE dropout} & \textbf{Ours} & \underline{98.23} & \underline{64.38} & \textbf{76.30} & 72.12 & \textbf{77.76} \\
\bottomrule
\end{tabular}}
\end{center}
\end{table}

\begin{table}[tb!]
\begin{center}
\caption{Comparison of performance (\%) among different methods using ResNet-18 on Office-Home \cite{venkateswara2017deep}.
The best and second-best are \textbf{bolded} and \underline{underlined}, respectively.
}
\label{table: OfficeHome}
\scalebox{1.0}{
\renewcommand\arraystretch{1.1}
\begin{tabular}
{l | c | cccc |c}
\toprule
Method & Venue & Artistic & Clipart & Product & Real & Avg.\\
\midrule
    RSC \cite{huang2020self} & ECCV 2020 & 57.70 & 48.58 & 72.59 & 74.17 & 63.26 \\
    MixStyle \cite{zhou2020domain} & ICLR 2021 & 58.70 & 53.40 & 74.20 & 75.90 & 65.50\\
    SagNet \cite{nam2021reducing} & CVPR 2021 & \underline{60.20} & 45.38 & 70.42 & 73.38 & 62.34 \\
    FACT \cite{xu2021fourier} & CVPR 2021 & \textbf{60.34} & 54.85 & \underline{74.48} & \underline{76.55} & \underline{66.56} \\
    StyleNeophile \cite{kang2022style} & CVPR 2022 & 59.55 & 55.01 & 73.57 & 75.52 & 65.89 \\
    COMEN \cite{chen2022compound} & CVPR 2022 & 57.60 & \underline{55.80} & \textbf{75.50} & \textbf{76.90} & 66.50 \\
    \midrule
    DeepAll \cite{zhou2020deep} {\scriptsize (\textit{our imple.})}  & AAAI 2020 & 52.06 & 46.12 & 70.45 & 72.45 & 60.27 \\
    \textbf{+ PLACE dropout} & \textbf{Ours} & \underline{60.20} & 54.02 & 73.80 & 76.13 & 66.04 \\ 
    \midrule
    DeepAll$^{++}$ & \textbf{Ours} & 56.83 & 55.49 & 70.12 & 72.28 & 63.68 \\
    \textbf{+ PLACE dropout} & \textbf{Ours} & 59.99 & \textbf{58.40} & 74.14 & 76.29 & \textbf{67.20} \\
\bottomrule
\end{tabular}}
\end{center}
\end{table}
\vspace{0.1cm}

\textbf{Results on OfficeHome.} We conducted experiments on OfficeHome and reported the results in Tab.~\ref{table: OfficeHome}.
OfficeHome is a more challenging benchmark for DG due to relatively smaller domain shifts and a larger number of categories compared to PACS and VLCS datasets.
Despite these challenges, our method still performs competitively with the latest DG methods on this benchmark, which improves the baseline performance by $5.77\%$ ($66.04\%$ vs. $60.27\%$) without extra computational cost.
We observe that PLACE dropout performs well on the Artistic, Product, and Realworld tasks, but achieves relatively mediocre results on the most difficult task, Clipart. The possible reason for this difference in performance is that the Clipart dataset contains more noise compared to the other three domains, which negatively impacts the model performance.
As shown in Tab.~\ref{table: OfficeHome}, some data augmentation methods can combat this noise and perform better on the Clipart domain compared to other methods, \textit{e.g.}, MixStyle \cite{zhou2020domain} and FACT \cite{xu2021fourier}. By combining PLACE dropout with the augmentation-based DeepAll$^{++}$, our method achieves a significant improvement over the baseline, \textit{i.e.}, $12.28\%$ ($58.40\%$ vs. $46.12\%$), and surpasses the nearest competitor COMEN \cite{chen2022compound} by $2.60\%$ ($58.40\%$ vs. $55.80\%$) on the Clipart dataset.
Furthermore, our method outperforms SOTA DG methods, \textit{e.g.}, StyleNeophile \cite{kang2022style} by $1.31\%$ ($67.20\%$ vs. $65.89\%$) and FACT \cite{xu2021fourier} by $0.64\%$ ($67.20\%$ vs. $66.56\%$).
The results further support the effectiveness of PLACE dropout in challenging DG tasks.

\begin{figure}[tb!]
    \centering
    \subfigure[Accuracy ($\%$) on DeepAll]{
    \begin{minipage}[t]{0.45\linewidth} 
    \label{figure: dropout methods baseline}
    \centering
    \includegraphics[scale=0.36]{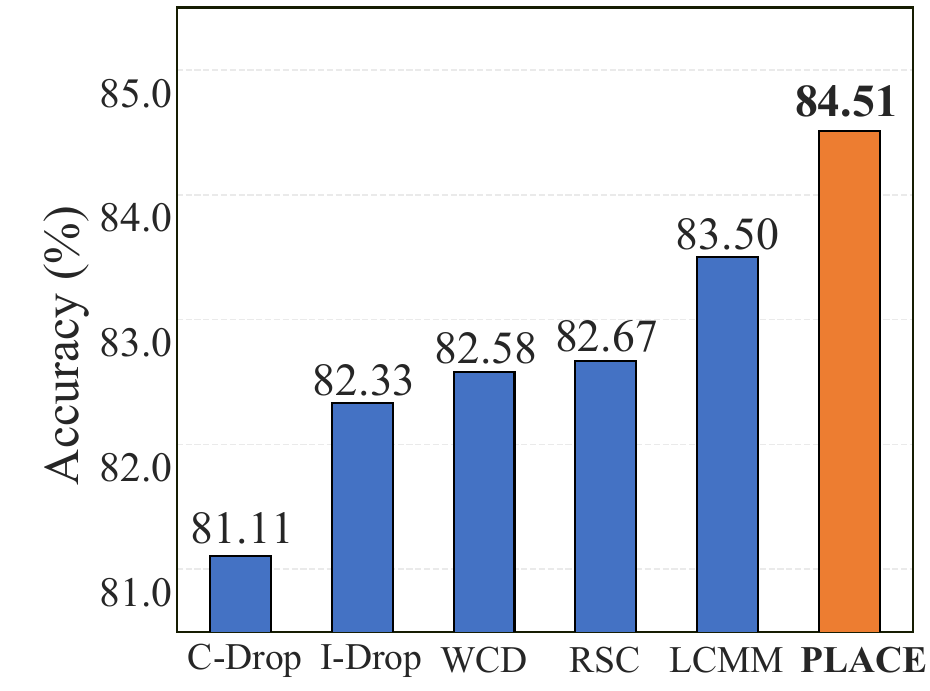}
    \end{minipage}
    }
    \subfigure[Accuracy ($\%$) on DeepAll$^{++}$]{
    \begin{minipage}[t]{0.45\linewidth} 
    \label{figure: dropout methods strong}
    \centering
    \includegraphics[scale=0.36]{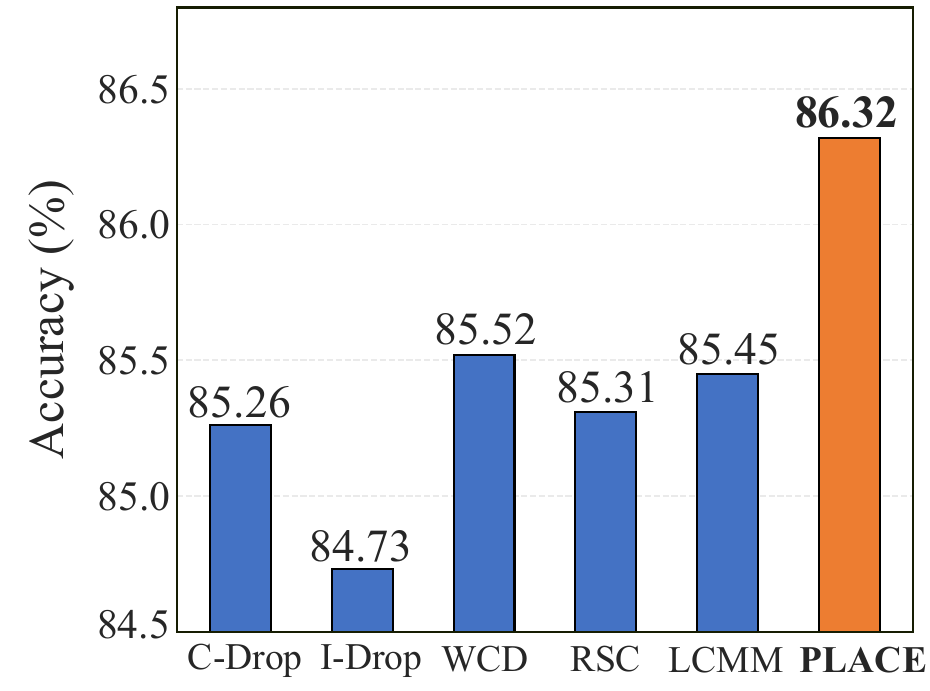}
    \end{minipage}
    }
    \vspace{-0.2cm}
    \caption{\textcolor{revisecolor}{Comparison of PLACE dropout and other dropout-based methods on PACS using ResNet-$18$.}
    }
    \label{figure: dropout methods}
    \vspace{-0.2cm}
\end{figure}
    
\begin{table}[tb!]
        \begin{center}
        \caption{\textcolor{revisecolor}{Effect (\%) of incorporating PLACE dropout on other SOTA methods on the PACS dataset. For the CNN-based methods, \textit{i.e.}, MixStyle \cite{zhou2020domain} and FACT \cite{xu2021fourier}, we use the ResNet-18 as the backbone. We also validate the effectiveness of PLACE dropout on two SOTA MLP-like networks, \textit{i.e.}, GFNet \cite{rao2021global} and ViP \cite{hou2022vision}.}}
        \label{table: SOTA + PLACE}
        \scalebox{1.0}{
        \begin{tabular}
        {l|cccc|c}
            \toprule
            Method & Art & Cartoon & Sketch & Photo & Avg.\\
            \midrule
            MixStyle \cite{zhou2020domain} {\scriptsize (ICLR'20)} & 84.10 & 78.80 & 75.90 & \textbf{96.10} & 83.70 \\
            \textbf{+ PLACE dropout} & \textbf{84.62} & \textbf{79.78} & \textbf{83.66} & 95.69 & \textbf{85.94} \\
            \midrule
            FACT \cite{xu2021fourier} {\scriptsize (CVPR'21)} & 85.37 & 78.38 & 79.15 & \textbf{95.15} & 84.51 \\
            \textbf{+ PLACE dropout} & \textbf{85.50} & \textbf{82.08} & \textbf{83.18} & 94.91 & \textbf{86.42} \\
            \midrule
            GFNet \cite{rao2021global} {\scriptsize (NeurIPS'21)} & 89.37 & 84.74 & 79.01 & 97.94 & 87.76\\
            \textbf{+ PLACE dropout} & \textbf{92.19} & \textbf{86.86} & \textbf{84.22} & \textbf{98.44} & \textbf{90.43} \\
            \midrule
            ViP \cite{hou2022vision} {\scriptsize (TPAMI'22)} & 88.09 & 84.22 & 82.41 & 98.38 & 88.27 \\
            \textbf{+ PLACE dropout} & \textbf{91.66} & \textbf{85.47} & \textbf{85.62} & \textbf{99.04} & \textbf{90.45} \\
            \bottomrule
        \end{tabular}}
        \end{center}
\end{table}

\textbf{Comparisons with Dropout-based Methods.}
To comprehensively prove the effectiveness of PLACE dropout, 
we compare it with the SOTA dropout-based methods,
including methods designed for domain generalization (RSC \cite{huang2020self} \footnotemark[3] and I-Drop \cite{shi2020informative}), supervised learning (C-Drop \cite{morerio2017curriculum} and WCD \cite{hou2019weighted}), and image compression (LMMD \cite{liu2023exploring}).
\footnotetext[3]{We rerun the official codes of RSC with the same hyperparameters as mentioned in \cite{huang2020self} but fail to reproduce the reported results, which may be due to the different hardware environment. This problem is also reported in \cite{xu2021fourier,nuriel2021permuted,wang2021embracing}. To be fair, we compare our method with RSC in our environment.}
We conduct comparative experiments on both DeepAll and DeepAll$^{++}$, and the results are presented in Fig.~\ref{figure: dropout methods}.
\textcolor{revisecolor}{
As shown in Fig.~\ref{figure: dropout methods baseline}, our PLACE dropout achieves a significant improvement of $1.01\%$ ($84.51\%$ vs. $83.50\%$) compared to the second-best approach LCMM, which utilizes a learnable vector to explicitly discard redundant channels.
This enhancement is attributed to our method's ability to generate diverse data variants across multiple layers, effectively mitigating overfitting on source domains and promoting model generalization.
Furthermore, in Fig.~\ref{figure: dropout methods strong}, when applied to DeepAll$^{++}$, our PLACE dropout still achieves SOTA performance and outperforms all other dropout-based methods, highlighting its superiority in promoting model generalization performance.
Besides, it is worth noting that our DeepAll$^{++}$ significantly benefits most of the other dropout-based methods, \textit{e.g.}, the performance of WCD \cite{hou2019weighted} is boosted by $2.94\%$ ($85.52\%$ vs. $82.58\%$) on DeepAll$^{++}$ compared to that on DeepAll. 
}
We also observe a performance decline of I-Drop on DeepAll$^{++}$, possibly due to that on the strong baseline, the regions with relatively less information could still contain important semantic information.
In conclusion, PLACE dropout surpasses the SOTA dropout-based methods of DG on both DeepAll and DeepAll$^{++}$, demonstrating the effectiveness of our method in improving the generalization ability of the model.
Additionally, we provide preliminary evidence that dropout-based methods can cooperate with other data augmentation techniques to enhance model performance in practical scenarios \cite{zhou2021survey},  which we hope will inspire future research in DG.

\textbf{Incorporating PLACE dropout with other SOTA DG methods.} 
\textcolor{revisecolor}{
To validate the generalization of PLACE dropout, we conducted experiments by integrating it with other SOTA DG methods, including CNN-based methods MixStyle \cite{zhou2020domain} and FACT \cite{xu2021fourier}, as well as MLP-like networks GFNet \cite{rao2021global} and ViP \cite{hou2022vision}.
As shown in Tab.~\ref{table: SOTA + PLACE}, when combined with the SOTA CNN-based methods, PLACE dropout consistently achieves substantial performance improvements, with a margin of $2.24\%$ ($85.94\%$ vs. $83.70\%$) for MixStyle \cite{zhou2020domain} and $1.91\%$ ($86.42\%$ vs. $84.51\%$) for FACT \cite{xu2021fourier}.
Furthermore, when applied to GFNet \cite{rao2021global}, a novel network exploring feature dependencies in frequency space, PLACE dropout effectively enhances the performance with an improvement of $2.67\%$ ($90.43\%$ vs. $87.76\%$). Besides, our method outperforms ViP \cite{hou2022vision}, which learns long-range dependencies in both height and width directions through linear projections, by $2.18\%$ ($90.45\%$ vs. $88.27\%$).
The results indicate that PLACE dropout complements DG methods effectively, confirming that mitigating overfitting on source domains can further enhance model generalization ability.
}



\subsection{Ablation Studies}

\begin{table}[tb!]
    \begin{center}
    \caption{Effect (\%) of each component in PLACE dropout on PACS with ResNet-18 or ResNet-50 as the backbone architecture. For simplicity, we denote the channel-wise dropout as C, the layer-wise dropout as L, and the progressive scheme as P.
    The best and second-best are \textbf{bolded} and \underline{underlined}, respectively.}
    \label{table: PLACE}
    \scalebox{1.0}{
    \begin{tabular}{c|ccc|cccc|c}
        \toprule
        Method & C & L & P & Art & Cartoon & Sketch & Photo & Avg.\\
        \midrule
        \multicolumn{9}{c}{DeepAll} \\
        \midrule
        Baseline & - & - & - & 80.19 & 77.19 & 73.48 & \textbf{95.71} & 81.64 \\
        \midrule
        Variant $1$ & \checkmark & - & - & 79.02 & 76.44 & 79.93 & 92.02 & 81.85 \\
        Variant $2$ & \checkmark & - & \checkmark &  80.34 &  78.07 & \underline{80.91} & 94.17 & 83.38\\
        Variant $3$ & - & \checkmark & - & 79.95 & 76.82 & 75.90 & 95.28 & 81.99 \\
        Variant $4$ & - & \checkmark & \checkmark & 81.03 & 77.18 & 76.04 & 95.25 & 82.38 \\
        Variant $5$ & \checkmark & \checkmark & - &  \underline{82.23} & \textbf{78.34} & 80.40 & 95.23 & \underline{84.05} \\
        \midrule
        PLACE dropout & \checkmark & \checkmark & \checkmark & \textbf{82.60} & \underline{78.33} & \textbf{81.47} & \underline{95.65} & \textbf{84.51} \\
        \midrule
        \multicolumn{9}{c}{DeepAll$^{++}$} \\
        \midrule
        Baseline & - & - & - & 83.64 & 78.28 & 81.41 & \underline{96.34} & 84.92 \\
        \midrule
        Variant $1$ & \checkmark & - & - & 81.71 & 76.01 & 82.42 & 94.33 & 83.62 \\
        Variant $2$ & \checkmark & - & \checkmark & 82.63 & 77.56 & 83.27 & 95.09 & 84.64 \\
        Variant $3$ & - & \checkmark & - & 83.69 & 77.94 & 82.92 & \textbf{96.65} & 85.30 \\
        Variant $4$ & - & \checkmark & \checkmark & 84.47 & \underline{79.25} & 83.22 & 96.11 & 85.76 \\
        Variant $5$ & \checkmark & \checkmark & - & \underline{84.57} & 79.18 & \textbf{84.16} & 96.23 & \underline{86.03} \\
        \midrule
        PLACE dropout & \checkmark & \checkmark & \checkmark & \textbf{85.40} & \textbf{79.69} & \underline{83.97} & 96.23 & \textbf{86.32} \\
        \bottomrule
    \end{tabular}}
    \end{center}
\end{table}

\textbf{Impact of different components.} 
We conducted experiments to investigate the contribution of each component in PLACE dropout on PACS using ResNet-$18$ as the backbone. The results are presented in Tab.~\ref{table: PLACE}, and each component has demonstrated significant importance in our method.
Starting with the baseline, Variant $1$ represents Channel-wise dropout, which applies dropout in the channel dimension to multiple layers simultaneously.
\textcolor{revisecolor}{
    It achieves a slight improvement on DeepAll but results in performance degradation on DeepAll$^{++}$, indicating that the simultaneous use of dropout in multiple layers may cause excessive information loss and hinder model training. 
    Nevertheless, the results also show that Channel-wise dropout can generate strong regularization effects, as evidenced by its notable improvement on Sketch.
    On the other hand, Layer-wise dropout (Variant $3$) involves randomly selecting a network layer (Layers 1-3) at each iteration to apply dropout in the pixel dimension instead of the channel dimension. Pure Layer-wise dropout exhibited a slight improvement of $0.35\%$ over the Baseline ($81.99\%$ vs. $81.64\%$), indicating the limited regularization strength of dropout in the pixel dimension.
    To address these issues, we developed Layer-wise and Channel-wise dropout (Variant $5$), which significantly improved performance by $2.41\%$ on DeepAll ($84.05\%$ vs. $81.64\%$) and by $1.11\%$ on DeepAll$^{++}$ ($86.03\%$ vs. $84.92\%$). The results prove that Layer-wise and Channel-wise dropout can effectively mitigate the detrimental effects of noise on model training while providing strong regularization to combat overfitting.
}
Besides, we extended Variants $1$, $3$, and $5$ to Variants $2$, $4$, and PLACE dropout with the progressive scheme to verify its effectiveness.
Based on Variant $1$, the progressive scheme achieved a relatively significant improvement on both DeepAll (by $1.47\%$) and DeepAll$^{++}$ (by $1.02\%$), owing to it can reduce the hindrance of multi-layers dropout to model convergence at the beginning of training.
On other variants, the progressive scheme also consistently improves performance by about $0.5\%$ on average, verifying its effectiveness for adequately mitigating overfitting.
\textcolor{revisecolor}{
Specifically, we observe that the Baseline achieves remarkable performance on Photo.
It could be due to the similarity between Photo and the pre-trained dataset ImageNet \cite{xu2021fourier, huang2020self}, which allows the model performance to easily saturate on this domain. 
However, when applying dropout during training, it would generate additional noise and lead to performance fluctuations, \textit{e.g.}, a slight performance decrease on Photo. 
The issue could be alleviated by adjusting the random seed or the dropout rate.
Nevertheless, our method enhances the accuracy on other challenging domains and achieves the best overall performance, proving that the three modules are indispensable for superior generalization ability.
}

\textbf{Different dismensions to conduct dropout.} 
We investigate different variants of PLACE dropout that apply dropout in various dimensions, including pixel-wise dropout (PLACE-E), spatial-wise dropout (PLACE-S), and channel-wise dropout (PLACE-C). 
We also conducted experiments to combine PLACE-S and PLACE-C (denoted as PLACE-S/C), 
which randomly selects either PLACE-S or PLACE-C to perform dropout with a probability of $50\%$ at each iteration.
As shown in Tab.~\ref{tab: different dimensions}, all the variants show performance gains, with PLACE-S and PLACE-C achieving larger improvements compared to PLACE-E.
The results suggest that structural dropouts, \textit{i.e.}, spatial- and channel-wise dropout, contribute to better resistance against overfitting.
Moreover, PLACE-C shows the best performance on both DeepAll and DeepAll$^{++}$, confirming the discussion in Section~\ref{section: channel-wise dropout} that channel-wise dropout can effectively perturb the patterns learned by the model, thereby providing a strong regularization effect.
\textcolor{revisecolor}{
Besides, we observed that the combined usage of PLACE-C and PLACE-S (PLACE-S\&C) yields improved results compared to using PLACE-S alone, but slightly reduces the accuracy compared to PLACE-C. 
A possible reason for the observation is that the regularization effects of PLACE-S and PLACE-C are not orthogonal, as both methods provide regularization by discarding a subset of feature maps.
Since numerous regions within the spatial dimension contain redundant information, as already discussed in Section~\ref{section: channel-wise dropout}, the regularization effect of PLACE-S could not be as effective as that of PLACE-C, thus slightly weakening the regularization effect when combined with PLACE-C. 
To effectively combat the overfitting on source domains, we apply PLACE dropout in the channel dimension to further improve the model generalization.
}

\begin{table}[tb!]
    \centering
    \caption{
        \textcolor{revisecolor}{Comparison of different variants of PLACE dropout that conduct dropout in different dimensions, \textit{i.e.}, pixel dimension (PLACE-E), spatial dimension (PLACE-S), channel dimension (PLACE-C), and spatial or channel dimension (PLACE-S\&C).
        The experiments are conducted on PACS using ResNet-18 as the backbone architecture.
        The best and second-best are \textbf{bolded} and \underline{underlined}, respectively.}}
    \scalebox{1.0}{
    \begin{tabular}{l|cccc|c}
        \toprule
        Method & Art & Cartoon & Sketch & Photo & Avg.\\
        \midrule
        DeepAll & 80.19 & 77.19 & 73.48 & \textbf{95.71} & 81.64 \\
        + PLACE-E & 81.20 & \underline{78.11} & 73.26 & 95.40 & 81.99 \\
        + PLACE-S & 82.08 & 77.90 & 78.18 & 95.07 & 83.31 \\
        \textbf{+ PLACE-C} & \textbf{82.60} & \textbf{78.33} & \textbf{81.47} & \underline{95.65} & \textbf{84.51} \\
        + PLACE-S/C & \underline{82.13} & 77.98 & \underline{78.95} & 95.39 & \underline{83.61} \\
        \midrule
        DeepAll$^{++}$ & 83.64 & 78.28 & 81.41 & \textbf{96.34} & 84.92 \\
        + PLACE-E & 83.77 & 79.02 & 82.51 & 96.15 & 85.36 \\
        + PLACE-S & 84.67 & 79.27 & 82.82 & 96.07 & 85.71 \\
        \textbf{+ PLACE-C} & \textbf{85.40} & \textbf{79.69} & \textbf{83.97} & \underline{96.23} & \textbf{86.32} \\
        + PLACE-S/C & \underline{84.97} & \underline{79.34} & \underline{83.38} & 96.13 & \underline{85.96} \\
        \bottomrule
    \end{tabular}}
    \label{tab: different dimensions}
\end{table}

\begin{table}[tb!]
    \begin{center}
    \caption{Effect (\%) of different layers where PLACE dropout is inserted on the PACS dataset with ResNet-$18$ or ResNet-$50$ as the backbone architecture.
    For notation, ${\rm L}1$ means PLACE dropout is applied after the first residual block; $\{{\rm L}1, {\rm L}2\}$ means PLACE dropout is applied after the first or the second blocks randomly. 
    The best and second-best are \textbf{bolded}
    and \underline{underlined}, respectively.
    }
    \label{table: Layers}
    \scalebox{1.0}{
    \begin{tabular}{c|cccc|cc}
    \toprule
        Method & \multicolumn{4}{c|}{Position} & \multicolumn{2}{c}{Baseline} \\
        \midrule
        & L1 & L2 & L3 & L4 & DeepAll & DeepAll$^{++}$\\
        \midrule
        Baseline & - & - & - & - & 81.64 $\pm$ 0.39 & 84.92 $\pm$ 0.16 \\
        \midrule
        Variant $1$ & \checkmark & - & - & - & 82.81 $\pm$ 0.03 & 85.67 $\pm$ 0.13 \\
        Variant $2$ & - & \checkmark & - & - & 83.63 $\pm$ 0.15 & 85.66 $\pm$ 0.04\\
        Variant $3$ & - & - & \checkmark & - & 83.25 $\pm$ 0.13 & 85.91 $\pm$ 0.12\\
        Variant $4$ & - & - & - & \checkmark & 81.48 $\pm$ 0.40 & 84.83 $\pm$ 0.34 \\
        \midrule
        Variant $5$ & \checkmark & \checkmark & - & - & 83.58 $\pm$ 0.15 & 85.69 $\pm$ 0.31 \\
        Variant $6$ & \checkmark & \checkmark & \checkmark & - & \textbf{84.51 $\pm$ 0.09} & \textbf{86.32 $\pm$ 0.15} \\
        Variant $7$ & \checkmark & \checkmark & \checkmark & \checkmark & \underline{83.85 $\pm$ 0.12} & \underline{86.06 $\pm$ 0.20} \\
    \bottomrule
    \end{tabular}}
    \end{center}
\end{table}

\textbf{Where to apply PLACE dropout.} 
To investigate which layers to apply PLACE dropout,
we conducted the experiments on PACS using the ResNet-$18$ architecture.
Given that a standard ResNet model has four residual blocks denoted by L$1$-$4$, we trained various models with PLACE dropout applied to different layers.
The results are presented in Tab.~\ref{table: Layers}.
We first tested the performance of the models with PLACE dropout in one single layer, \textit{i.e.}, Variant $1$-$4$.
The model with PLACE dropout in the first three layers achieves consistent improvement, indicating that PLACE dropout can alleviate the overfitting issue in every network layer.
However, inserting dropout in the $4$-th layer reduces the model accuracy on both DeepAll and DeepAll$^{++}$, which might be due to ${\rm L4}$ being near the prediction layer and tending to capture label-related information, thus randomly dropout is likely to break the inherent label space and hamper the model generalization ability.
We further tested the model performance with PLACE dropout in multiple layers.
By comparing Variant $6$ with $7$, we observed that dropout in the $4$-th layer reduces the performance, which is likely related to the randomness of dropout and could be overcome by carefully guiding the dropout \cite{hou2019weighted,huang2020self}.
Finally, Variant $6$ with PLACE dropout in the $1$-st, $2$-nd, and $3$-rd layers achieves the best performance and yields the baseline $2.87\%$ ($84.51\%$ vs. $81.64\%$) on DeepAll and $1.14\%$ ($86.06\%$ vs. $84.92\%$) on DeepAll$^{++}$, aligning well with the discussion for the layer-wise dropout in Section~\ref{layer-wise dropout}.

\textbf{Sensitivity to the maximum dropout rate.} 
To investigate the optimal value of the hyperparameter ${\rm P_{max}}$, which indicates the maximum proportion of dropping channels, we conducted the experiments on PACS with ResNet-$18$ as the backbone.
The experimental results on both DeepAll and DeepAll$^{++}$ are presented in Tab. \ref{table: Pmax PACS}.
Remarkably, the model performance remains relatively consistent across different values of ${\rm P_{max}}$, and it achieves excellent performance when ${\rm P_{max}}$ is set between $25\%$ and $40\%$ on both baselines. Particularly, our method achieves the highest average accuracy on both DeepAll and DeepAll$^{++}$ when ${\rm P_{max}}$ is set to $33\%$, indicating that the baseline does not significantly influence the selection of ${\rm P_{max}}$.
Considering that the maximum dropout rate ${\rm P_{max}}$ influences the strength of the regularization effect during training, the optimal value should not be excessively large or small. A too large ${\rm P_{max}}$ could lead to excessive information loss during training, hindering the model from learning discriminative information from source domains. Conversely, a too small ${\rm P_{max}}$ might fail to effectively address the overfitting issue on source domains.
Based on the experimental results, We adopt $33\%$ as the default setting in all experiments if not specified.

\begin{table}[tb!]
    \begin{center}
    \caption{Effect($\%$) of PLACE dropout with different ${\rm P_{max}}$ to model performance on the PACS dataset with ResNet-$18$ as the backbone architecture.
    The best and second-best are \textbf{bolded} and \underline{underlined}, respectively.
    }
    \label{table: Pmax PACS}
    \scalebox{1.0}{
    \begin{tabular}
    {l|c|cccc|c}
        \toprule
        & ${\rm P_{max}}$ & Art & Cartoon & Sketch & Photo & Avg.\\
        \midrule
        \multirow{4}{*}{DeepAll} & $20\%$ & 81.51 & 77.68 & 78.71 & \textbf{95.67} & 83.39 \\
        & $25\%$ & \underline{82.31} & 78.09 & 79.98 & 95.64 & 84.01 \\
        & $\textbf{33\%}$ & \textbf{82.60} & \textbf{78.33} & \textbf{81.47} & \underline{95.65} & \textbf{84.51}\\
        & $40\%$ & 82.14 & \underline{78.27} & \underline{80.81} & 95.29 & \underline{84.13} \\
        \midrule
        \multirow{4}{*}{DeepAll$^{++}$} & $20\%$ & 83.94 & 79.14 & 83.13 & \underline{96.71} & 85.73 \\
        & $25\%$ & \underline{84.63} & \underline{79.89} & 82.85 & 96.70 & 86.02\\
        & $\textbf{33\%}$ & \textbf{85.40} & 79.69 & \textbf{83.97} & 96.23 & \textbf{86.32}\\
        & $40\%$ & 83.94 & \textbf{79.91} & \underline{83.56} & \textbf{96.77} & \underline{86.05} \\
        \bottomrule
    \end{tabular}}
    \end{center}
\end{table}

\subsection{Further analysis}

\begin{figure}[b!]
\centering
\vspace{-0.2cm}
    \subfigure[$d_{drop}^{i} - d_{bal}^{i}$ on DeepAll.]{
    \begin{minipage}[b]{0.45\linewidth}
    \centering
        \includegraphics[scale=0.34]{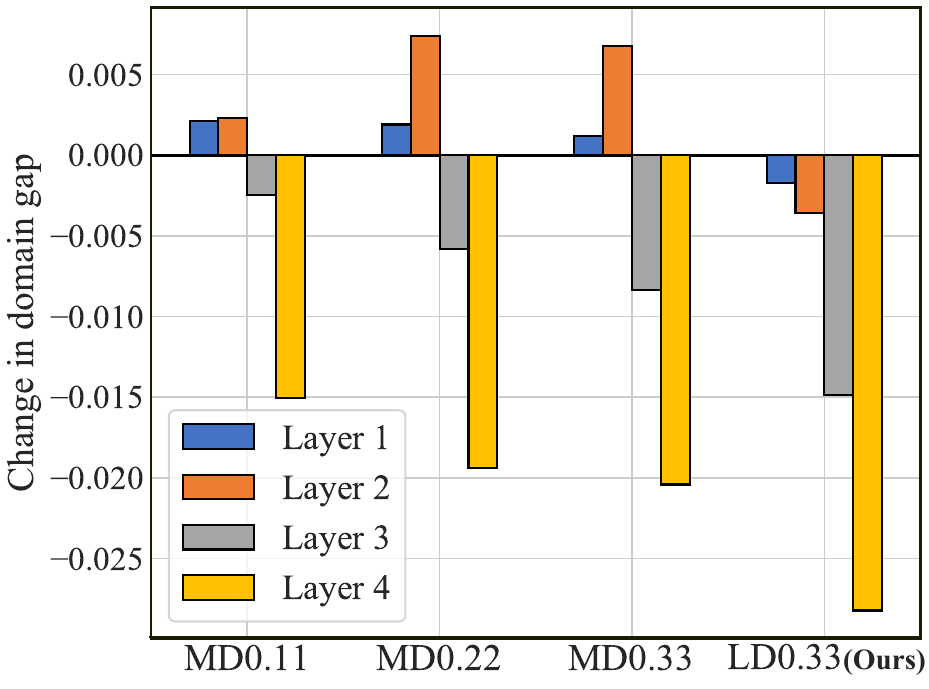}
    \end{minipage}
    }
    \subfigure[$d_{drop}^{i} - d_{bal}^{i}$ on DeepAll$^{++}$.]{
    \begin{minipage}[b]{0.45\linewidth}
    \centering
        \includegraphics[scale=0.34]{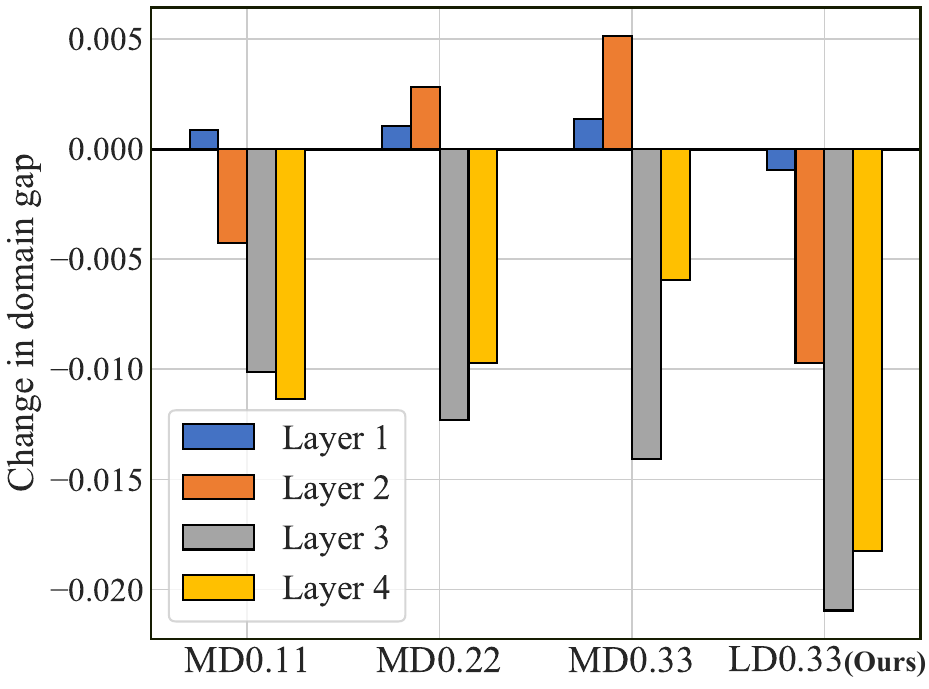}
    \end{minipage}
    }
    \vspace{-0.3cm}
    \caption{
        \textcolor{revisecolor}{
        Comparison of inter-domain discrepancy between multi-layers dropout and our layer-wise dropout. 
        We denote multi-layers dropout as ${\rm MD}$ and layer-wise dropout as ${\rm LD}$, \textit{e.g.}, ${\rm MD}0.11$ is multi-layers dropout with drop ratio $p$ as $0.11$.
        (a) and (b) are the results on DeepAll and DeepAll$^{++}$, respectively.
        The smaller $d_{drop}^{i} - d_{bal}^{i}$ indicates the larger reduction of the domain gap and the better generalization of the model.
        }
    }
    \label{figure: domaingap layerwise}
    \vspace{-0.4cm}
\end{figure}

\begin{table}[tb!]
    \begin{center}
    \caption{Comparison of layer-wise dropout and multi-layers dropout on PACS with ResNet-$18$ (Best in \textbf{bold}). For simplicity, We denote multi-layers dropout as ${\rm MD}$ and layer-wise dropout as ${\rm LD}$, \textit{e.g.}, ${\rm MD}0.11$ is multi-layers dropout with ${\rm P_{max}}$ as $0.11$. The best and second-best are \textbf{bolded} and \underline{underlined}, respectively.}
    \label{table: MD}
    \scalebox{1.0}{
    \begin{tabular}{c | c | cccc |c}
    \toprule
    & Method & Art & Cartoon & Sketch & Photo & Avg. \\
    \midrule
    \multirow{4}{*}{DeepAll} & MD$0.11$ & 81.91 & 78.08 & 78.86 & \underline{95.64} & 83.62 \\
    & MD$0.22$ & \underline{82.42} & \underline{78.44} & 80.27 & 94.37 & \underline{83.88} \\
    & MD$0.33$ & 80.25 & \textbf{78.50} & \textbf{81.64} & 94.23 & 83.65 \\
    & \textbf{LD$\textbf{0.33}$} & \textbf{82.60} & 78.33 & \underline{81.47} & \textbf{95.65} & \textbf{84.51} \\
    \midrule
    \multirow{4}{*}{DeepAll$^{++}$} & MD$0.11$ & 84.18 & \underline{78.75} & 83.33 & \underline{95.93} & 85.55 \\
    & MD$0.22$ & \underline{84.25} & 78.63 & \textbf{84.28} & 95.91 & \underline{85.77} \\
    & MD$0.33$ & 82.45 & 77.22 & 83.33 & 94.97 & 84.49 \\
    & \textbf{LD$\textbf{0.33}$} & \textbf{85.40} & \textbf{79.69} & \underline{83.97} & \textbf{96.23} & \textbf{86.32} \\
    \bottomrule
    \end{tabular}}
    \end{center}
\end{table}

\textbf{Layer-wise dropout helps reduce domain gap.} 
To analyze the effect of layer-wise dropout on the model generalization, we conducted experiments to measure the inter-domain discrepancy of extracted features from intermediate layers with PLACE dropout on both DeepAll and DeepAll$^{++}$. The discrepancy was computed using Eq. (\ref{eq:inter-domain instance}). We also compared Layer-wise dropout (LD) with Multi-layers dropout (MD), where dropout with lower dropping rates is simultaneously applied to ${\rm L}1-3$ of the model.
As shown in Fig.~\ref{figure: domaingap layerwise}, layer-wise dropout effectively reduces the inter-domain distances across all network layers, leading to a significant improvement in model generalization, regardless of whether applied to DeepAll or DeepAll$^{++}$.
\textcolor{revisecolor}{
    Specifically, when we employ MD instead of LD, we observe that the values of $d_{drop}^i - d_{bal}^i$ in the $2$-th layer are positive, indicating an increase in inter-domain discrepancy on that specific layer. This aligns well with the discussion in Sec.~\ref{layer-wise dropout}, indicating that simultaneous dropout across multiple layers could lead to cumulative noise that hinders some layers from fully learning, causing an increase in the inter-domain gap on certain layers, \textit{e.g.}, the $2$-th layer. 
    To tackle this issue, we design the Layer-wise dropout, where we randomly select one layer to adopt dropout at each iteration, thereby mitigating excessive noise that could impede the learning process.
    As demonstrated in "LD0.33" of Fig.~\ref{figure: domaingap layerwise}, Layer-wise dropout achieves a smaller domain gap than multi-layers dropout, especially for the $2$-th layer, indicating that it can narrow the domain gap between source and target domains more effectively than simply utilizing dropout for multiple layers with a lower dropping rate.
}
Furthermore, we provide detailed comparison results on both DeepAll and DeepAll$^{++}$ in Tab.~\ref{table: MD}. The model with layer-wise dropout consistently achieves superior performance compared to the model with multi-layers dropout. These results verify that simply utilizing dropout for multiple layers with a lower dropping rate cannot lead to the same regularization as our layer-wise dropout. Additionally, the results demonstrate that layer-wise dropout can effectively reduce the domain gap between source and target domains, thereby improving the model generalization ability.

\begin{figure}[tb!]
\centering
\subfigure[Accuracy on the train set.]{
    \begin{minipage}[t]{0.23\linewidth}
    \centering
        \includegraphics[scale=0.22]{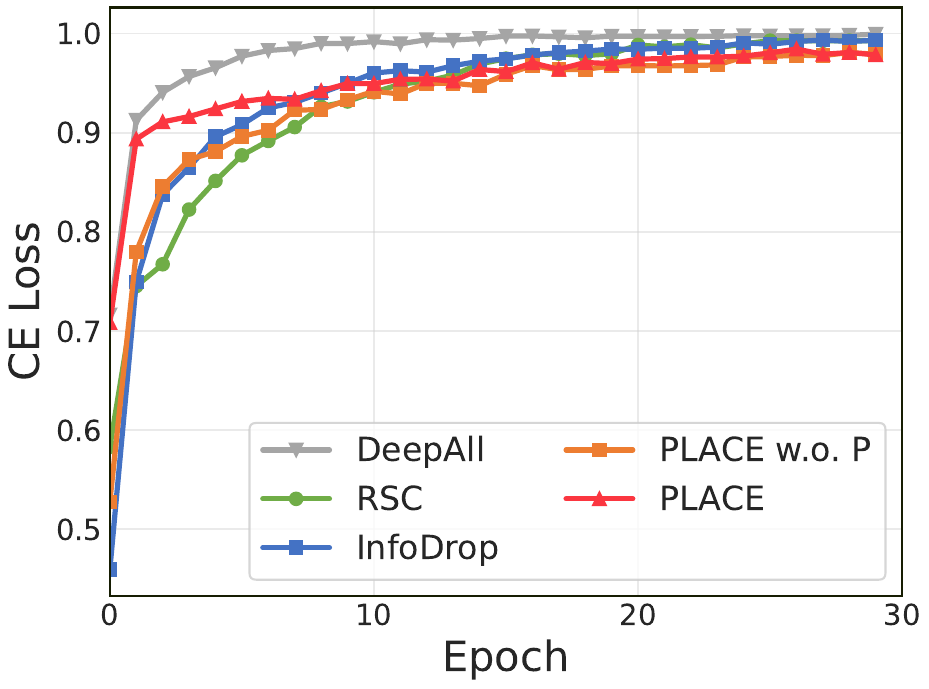} 
    \end{minipage}
    \label{figure:bound train acc}
}
\subfigure[Accuracy on the test set.]{
    \begin{minipage}[t]{0.23\linewidth} 
    \centering
        \includegraphics[scale=0.22]{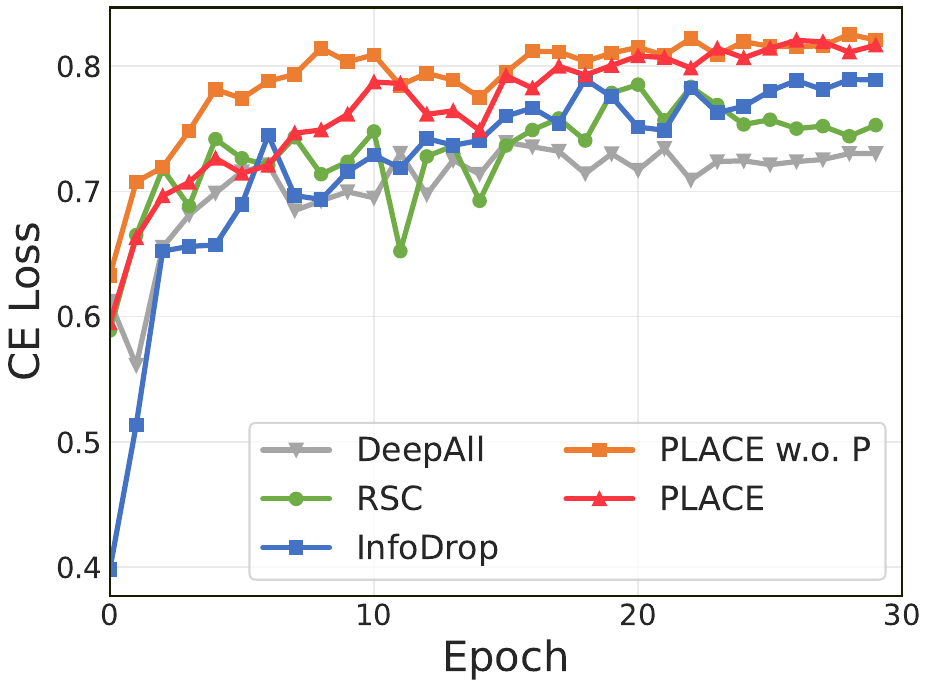}
    \end{minipage}
    \label{figure:bound test acc}
}
\subfigure[Loss on the test set.]{
    \begin{minipage}[t]{0.23\linewidth}
    \centering
        \includegraphics[scale=0.22]{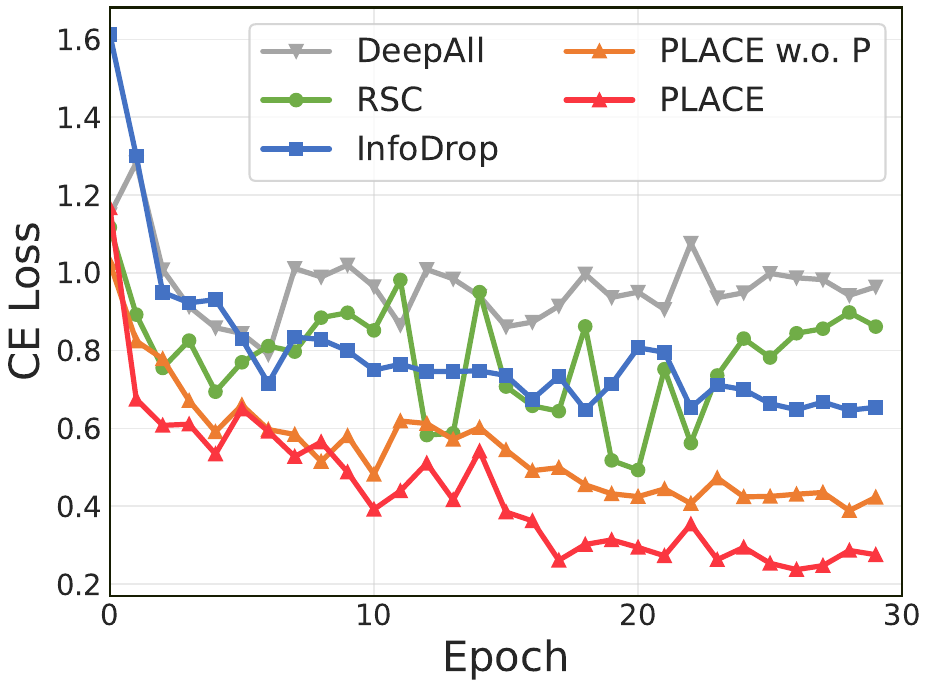}
    \end{minipage}
    \label{figure:bound test loss}
}
\subfigure[Loss difference $\widehat{\xi}_t(h)$.]{
    \begin{minipage}[t]{0.23\linewidth} 
    \centering
        \includegraphics[scale=0.22]{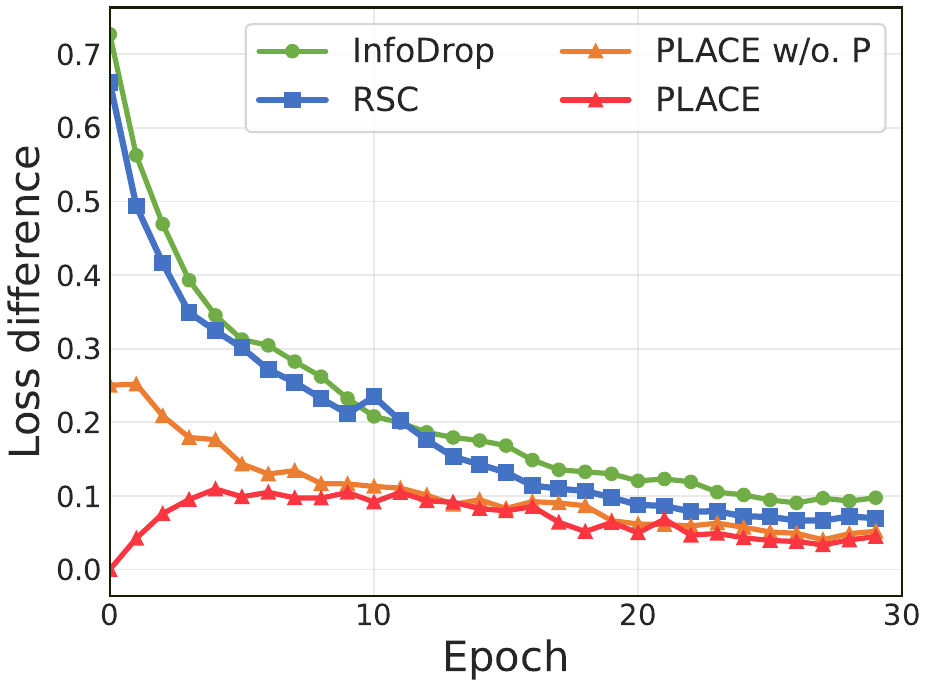}
    \end{minipage}
    \label{figure:bound loss difference}
}
\vspace{-0.2cm}
\caption{The influence of PLACE dropout on the procedure of model training and test. We compare PLACE dropout with the baseline DeepAll and RSC \cite{huang2020self}. We also investigate the effect of the progressive scheme. (a) presents the accuracy of each method on the train set, (b) and (c) are the accuracy and loss of each method on the test set, and (d) shows the change of loss difference $\xi_t(h)$ as the training progresses.}
\label{figure: bound}
\vspace{-0.1cm}
\end{figure}

\begin{table}[tb!]
    \centering
    \caption{The distribution discrepancy ($\times 10$) of inter-domain (across all source domains) and intra-class (for all classes in source domains) on PACS. Note that the lower the value indicates the better performance.}
    \vspace{-0.2cm}
    \scalebox{1.0}{
    \begin{tabular}{l | cc | cc | cc}
      \toprule
      Method & \multicolumn{2}{c}{PACS} & \multicolumn{2}{c}{VLCS} & \multicolumn{2}{c}{OfficeHome} \\
      \midrule
       & Inter-domain & Intra-class & Inter-domain & Intra-class & Inter-domain & Intra-class \\
      \midrule
      DeepAll & 2.43 & 2.95 & 1.58 & 2.41 & 0.76 & 1.49\\
      + PLACE & \textbf{1.96} & \textbf{2.48} & \textbf{1.54} & \textbf{2.35} & \textbf{0.71} & \textbf{1.44}\\
      \midrule
      DeepAll$^{++}$ & 1.63 & 2.13 & 1.47 & 2.34 & 0.69 & 1.40 \\
      + PLACE & \textbf{1.53} & \textbf{2.04} & \textbf{1.38} & \textbf{2.27} & \textbf{0.63} & \textbf{1.36}\\
      \midrule
      \end{tabular}}
    \label{tab:distance}
  \end{table}

\textbf{PLACE dropout indicates tighter generalization bound.} 
\label{discuss bound}
We investigate the effect of PLACE dropout on the model generalization bound. 
As is shown in Fig.~\ref{figure:bound train acc}, InfoDrop, RSC, and PLACE dropout can all increase the difficulty of model training, but layer-wise and channel-wise dropout exhibits the strongest regularization effect in the late stages of trainin, and the progressive scheme ensures stable model convergence in the early training stages.
Fig.\ref{figure:bound test acc} and Fig.\ref{figure:bound test loss} illustrate the accuracy and loss on the test set with an increasing number of epochs for each method. 
Our proposed method consistently reduces the model prediction loss and enhances accuracy more effectively on the test set compared to other methods. 
Furthermore, the progressive scheme, when combined with layer-wise and channel-wise dropout, further improves model performance by protecting early-stage convergence.
By employing the progressive scheme, the model learns "smarter" weights, avoiding learning potentially harmful information \cite{morerio2017curriculum} and effectively addressing overfitting. 
Fig.~\ref{figure:bound loss difference} presents the changes of $\widehat{\xi}_t(h)$ during training, reflecting the evolution of the generalization error bound.
PLACE dropout enhances the model robustness against perturbations caused by dropout, \textit{i.e.}, reducing the empirical loss difference $\widehat{\xi}_t(h)$, which aligns well with the theoretical discussion that layer-wise and channel-wise dropout can yield a tight generalization error bound. The empirical results also validate the effectiveness of the progressive scheme.

\textbf{Distribution Discrepancy of Feature Representations}
To investigate the influence of our PLACE dropout, we examine the inter-domain (across all source domains) and intra-class (inter-domain distance for the same-class samples) discrepancy.
The inter-domain discrepancy and the intra-class discrepancy are computed by the following formulations:
\begin{equation}
    d_{inter-domain} = \frac{2}{K(K-1)} \sum_{m=1}^{K} \sum_{n=m+1}^{K} || \overline{\textbf{z}}_{m} - \overline{\textbf{z}}_{n} ||_2,
\end{equation}
\begin{equation}
    d_{intra\mbox{-}class} = \frac{1}{C K} \sum_{c=1}^{C} \sum_{k=1}^{K}||\overline{\textbf{z}}_{k, c} - \overline{\textbf{z}}_{k}||_2,
\end{equation}
where $\overline{\textbf{z}}_{k}$, $\overline{\textbf{z}}_{m}$, $\overline{\textbf{z}}_{n}$ are the feature means of all classes in the $k$-th, $m$-th, $n$-th source domain, respectively. $\overline{\textbf{z}}_{k, c}$ is the averaged representation of the $c$-th class in the $k$-th source domain, $K$ denotes the number of source domains and $C$ is the number of categories. 
As shown in Tab. \ref{tab:distance}, our method can effectively reduce the domain gap of source domains on both DeepAll and DeepAll$^{++}$, especially on PACS which shows a large domain gap.
Even on the more challenging datasets VLCS and OfficeHome, our method can still effectively close the domain gap.
Besides, PLACE dropout can also reduce intra-class discrepancy, which indicates that it can effectively capture discriminative features.
The results demonstrate the effectiveness of our method in mitigating the overfitting issue, which is beneficial to reducing the intra-class and inter-domain distances among source domains.

\section{Conclusions}
In this paper, we propose a simple yet effective dropout-based method for domain generalization to mitigate the overfitting problem on source domains and help the model generalize well to unseen target domains. 
We name our method Progressive Layer-wise and Channel-wise (PLACE) dropout.
At each iteration, PLACE dropout randomly selects one middle network layer and then randomly mutes its channels by a proportion that gradually increases as the training progresses.
We also provide theoretical analysis for layer-wise and channel-wise dropout and prove its effectiveness in reducing the generalization error bound.
Extensive experiments on various datasets demonstrate that our method outperforms other related SOTA DG methods with no extra network parameters and almost no increment in computing cost.
\textcolor{revisecolor}{In future work, we will further analyze the overfitting problem in the DG task and develop advanced methods that enable the model to adaptively learn suitable layers and dropout rates based on input data.}

\begin{acks}
    The work is supported by NSFC Program (62222604, 62206052, 62192783) and Jiangsu Natural Science Foundation Project (BK20210224).
\end{acks}


\bibliographystyle{ACM-Reference-Format}

\bibliography{egbib}

\clearpage
\appendix

\section{Proof of Theoretical Analysis}
\subsection{Proof of Proposition 1}
\noindent\textbf{Proposition 1.} \textit{
    Let $\widehat{\xi}_t(h)$ be the estimated value of $\xi(h)$ at iteration $t$, which is defined as: $\widehat{\xi}_t(h) = |L(\textbf{z}_t, h_t) -  L(\widetilde{\textbf{z}}_t, h_t)|$. 
    Given a sufficiently small learning rate $\eta$, if discarding structural features will increase the empirical loss at the current iteration $t$, 
    it holds that the layer-wise and channel-wise dropout can continually decrease $\widehat{\xi}_t(h)$ and lead it to be a small number at the end of training.
}
\vspace{0.1cm}


\noindent\textit{\textbf{Proof.}} To investigate the trend of $\widehat{\xi}_{t}(h)$ over iterations, we first utilize the Taylor series formula to derive the expansion of the empirical loss $L(\textbf{z}_{t + 1}, h)$ of the model at iteration $t + 1$:
\begin{equation}
    L(\textbf{z}_{t+1}, h_{t+1}) = L(\textbf{z}_t, h_{t}) + \frac{\partial  L(\textbf{z}_t, h_{t})}{\partial h_{t}} (h_{t+1} - h_{t}) + 
    \frac{1}{2}\frac{\partial^2  L(\textbf{z}_t, h_{t})}{\partial^2 h_{t}} ||h_{t+1} - h_{t}||^2 + \ldots.
    \label{eq:loss Taylor}
\end{equation}
Note that the update step size of the hypothesis $h_{t+1}$ can be denoted as: 
\begin{equation}
    h_{t+1} - h_{t} = -\eta\widetilde{\textbf{g}}_t, \hspace{0.5em} where \hspace{0.5em} \widetilde{\textbf{g}}_t = \frac{\partial  L(\widetilde{\textbf{z}_t}, h_{t})}{\partial h_{t}}.
\end{equation}
The assumption that discarding structural features will
increase the empirical loss could be formulated as: $L(\widetilde{\textbf{z}}_t, h_t) = \gamma_t L(\textbf{z}_t, h_t)$, $\gamma_t \geq 1$. 
The small learning rate $\eta$ allows us to discard the higher order terms in the Taylor expansion of empirical loss $L(\textbf{z}_{t + 1}, h)$ in Eq. (\ref{eq:loss Taylor}), then we can obtain:
\begin{equation}
    L(\textbf{z}_{t+1}, h_{t+1}) 
    = L(\textbf{z}_t, h_{t}) - \frac{1}{\gamma_t} \eta \widetilde{\textbf{g}_t}^2.
    \label{eq:gradient loss}
\end{equation}
We can also get an approximation of $L(\widetilde{\textbf{z}}_{t+1}, h_{t+1})$ through a similar process as above, which is formulated as $L(\widetilde{\textbf{z}}_{t+1}, h_{t+1}) = L(\widetilde{\textbf{z}}_{t}, h_{t}) - \eta \widetilde{\textbf{g}_t}^2$. Finally, we can derive the update step size of $\widehat{\xi}_{t}(h)$:
\begin{equation}
    \begin{aligned}
        \widehat{\xi}_{t+1}(h) - \widehat{\xi}_{t}(h)
        &= |L(\widetilde{\textbf{z}}_{t+1}, h_{t+1}) - L(\textbf{z}_{t+1}, h_{t+1})| - |L(\widetilde{\textbf{z}}_t, h_t) - L(\textbf{z}_t, h_t)| \\
        &= L(\widetilde{\textbf{z}}_{t+1}, h_{t+1}) - L(\widetilde{\textbf{z}}_t, h_t) - (L(\textbf{z}_{t+1}, h_{t+1}) - L(\textbf{z}_t, h_t)) \\
        &= - \eta \widetilde{\textbf{g}_t}^2 - ( - \frac{1}{\gamma_t} \eta \widetilde{\textbf{g}_t}^2) =-(1-\frac{1}{\gamma_t}) \eta \widetilde{\textbf{g}_t}^2 < 0.
    \end{aligned}
\end{equation}

The result indicates that layer-wise and channel-wise dropout can decrease $\widehat{\xi}_t(h)$ at every iteration, thus forcing it to be a small number at the end of training.
This discussion is also experimentally verified in Section 5.5 of the paper.
Therefore, layer-wise and channel-wise dropout can reduce the sensitivity of the model to dropout and produce a tight generalization error bound. $\hfill \blacksquare$

\subsection{Proof of Proposition 2}
\noindent\textbf{Proposition 2.} \textit{
Dropout in different layers perform as different data augmentations in the input space, thus the layer-wise dropout can generate more diverse augmented data than the single-layer dropout, \textit{i.e.}, increasing the dataset size $N$.
}
\vspace{0.1cm}

\noindent\textit{\textbf{Proof.}} Consider the hypothesis $h$ that is a network with $n$ hidden layers, and the $i$-th hidden layer is denoted as $l_i(\cdot)$.
Given the input image $x$, we denote the output of the $i$-th hidden layer as $f_i(x)$, and use $\widetilde{f}_i(x)$ to denote its perturbed version generated by dropout.
Let $m^{(i)}$ be the dropout mask in the $i$-th hinden layer, we can formulated the relationship between $\widetilde{f}_i(x)$ and $f_i(x)$ as:
\begin{equation}
    \widetilde{f}_i (x) = m_i \odot (f_i (x)).
    \label{eq:fx}
\end{equation}

Assuming that for a given $\widetilde{f}_i (x)$, we can find a $x_i'$ in the input space that satisfys $\widetilde{f}_i (x) = f_i(x_i')$. 
Then we will prove that for an input $x$, it is hard to find a single $x_i' = x_j'$ if $i \neq j$, where $\widetilde{f}_i (x) = f_i(x_i')$ and $\widetilde{f}_j (x) = f_j(x_j')$. For convenience, we consider the situation when $j \textless i$. From Eq. (\ref{eq:fx}), we have:
\begin{equation}
    \begin{aligned}
        \widetilde{f}_i(x) &= (m_i\odot l_i) \ldots (m_{j+1}\odot l_{j+1}) (\widetilde{f}_{j} (x)) \\
        &= (m_i\odot l_i) \ldots (m_{j+1}\odot l_{j+1}) (f_{j} (x_j')), \\
        f_i(x_i') &= l_i \circ l_{i-1} \ldots \circ l_{j+1} (f_{j} (x_i')).
    \end{aligned}
    \label{eq:perturb f}
\end{equation}

If there exist $x_i' = x_j'$, from Eq. (\ref{eq:perturb f}) we can derive that $\{m_i, m_{i-1}, \ldots, m_{j+1}\}$ does not apply any modification to $f_j(x_j')$, which means $m=1$ for $m \in \{m_i, m_{i-1}, \ldots, m_{j+1}\}$ when $f_j(x_j') > 0$. 
This situation is unlikely to happen unless the dropout rate $p = 0$ for all hidden layers between $l_i$ and $l_j$.
Therefore, inserting dropout into different layers performs as different data augmentations in the input space, which can generate diverse variants of data during training.
Based on this conclusion, our layer-wise and channel-wise dropout can effectively increase the dataset size $N$, leading to a tighter generalization error bound than single-layer dropout. $\hfill \blacksquare$

\section{Analysis for the Strong Baseline DeepAll$^{++}$}

\begin{table}[b!]
    \begin{center}
    \caption{Effect (\%) of PLACE dropout on different variants of DeepAll$^{++}$ on the PACS dataset with ResNet-18 as the backbone network. We denote SwapStyle as S and RandAug as R for simplicity.
    The best is \textbf{bolded}.}
    \label{table: DeepAll++ PACS}
    \scalebox{1.0}{
    \begin{tabular}
    {l|cccc|c}
        \toprule
        Method & Art & Cartoon & Sketch & Photo & Avg.\\
        \midrule
        DeepAll & 80.19 & 77.19 & 73.48 & 95.71 & 81.64 \\
        + PLACE dropout & 82.60 & 78.33 & 81.47 & 95.65 & 84.51\\
        \midrule
        DeepAll w. S & 81.59 & 78.03 & 76.43 & \textbf{96.71} & 83.19\\
        + PLACE dropout & 82.86 & 79.43 & 82.18 & 95.51 & 85.00 \\
        \midrule
        DeepAll w. R & 82.90 & 76.89 & 78.55 & 96.17 & 83.63\\
        + PLACE dropout & 84.14 & \textbf{79.89} & 79.95 & 96.11 & 85.02 \\
        \midrule
        DeepAll$^{++}$ & 83.64 & 78.28 & 81.41 & 96.34 & 84.92 \\
        + PLACE dropout & \textbf{85.40} & 79.69 & \textbf{83.97} & 96.23 & \textbf{86.32} \\
        \bottomrule
    \end{tabular}}
    \end{center}
\end{table}

\subsection{Effect of each components in DeepAll$^{++}$}
We conduct ablation studies for each component in DeepAll$^{++}$ on PACS with ResNet-$18$ as the backbone.
The experimental results are represented in Tab. \ref{table: DeepAll++ PACS}. 
The DeepAll$^{++}$ is a strong baseline model for domain generalization, which consists of two augmentation methods, \textit{i.e.}, the image-level augmentation method RandAug \cite{cubuk2020randaugment} and the feature-level augmentation method SwapStyle.
Specifically, SwapStyle can augment the data at the feature level by exchanging the feature statistics of any two images as the statistics can indicate the representative style information \cite{huang2017arbitrary,zhou2020domain}.
RandAug \cite{cubuk2020randaugment} is an effective automated data augmentation method to increase image-level diversity.
As shown in Tab.~\ref{table: DeepAll++ PACS}, both Swapstyle and RandAug can effectively improve the model performance.
SwapStyle improves the average accuracy performance of DeepAll by $1.55\%$ ($83.19\%$ vs. $81.64\%$) while RandAug improves by $1.99\%$ ($83.63\%$ vs. $81.64\%$).
As we can observe, Swapstyle can effectively improve the model performance on Cartoon and Photo, while RandAug is more effective on Art and Sketch.
By combining DeepAll with SwapStyle and RandAug, DeepAll$^{++}$ achieves the highest performance and outperforms DeepAll by $3.28\%$ ($84.92\%$ vs. $81.64\%$), indicating that the two augmentation methods are complementary with each other for improving model generalization.
We also test the performance of PLACE dropout on these different variants of DeepAll$^{++}$ and present the experimental results in Tab.~\ref{table: DeepAll++ PACS}.
We observe that the performance of PLACE dropout on Photo drops a little, which may be due to the task being saturated in performance.
Nevertheless, PLACE dropout can achieve significant improvements on all variants, which verifies its effectiveness and stability on different baselines.

\begin{table}[tb!]
    \centering
    \caption{Comparison of PLACE dropout with other start-or-the-art DG methods on DeepAll$^{++}$ using ResNet-18 as the backbone network. The experiments are conducted on the PACS dataset. The best is \textbf{bolded}.}
    \scalebox{1.0}{
    \begin{tabular}{l|cccc|c}
      \toprule
      Method & Art & Cartoon & Sketch & Photo & Avg.\\
      \midrule
      MMLD \cite{matsuura2020domain} & 81.28 & 77.16 & 72.29 & 96.09 & 81.83 \\
      +DeepAll$^{++}$ & 83.40 & 79.56 & 83.15 & 94.19 & 85.08 \\
      \midrule
      SagNet \cite{nam2021reducing} & 81.01 & 77.47 & 77.32 & 95.99 & 82.95 \\
      +DeepAll$^{++}$ & 83.89 & 79.67 & 80.63 & \textbf{96.41} & 85.15 \\
      \midrule
      PLACE (Ours) & 82.60 & 78.33 & 81.47 & 95.65 & 84.51 \\
      +DeepAll$^{++}$ & \textbf{85.40} & \textbf{79.69} & \textbf{83.97} & 96.23 & \textbf{86.32} \\
      \bottomrule
    \end{tabular}}
    \label{tab: DG SOTA methods}
  \end{table}

\subsection{Incorporating DeepAll$^{++}$ with other DG methods}
Besides incorporating the strong baseline DeepAll$^{++}$ with our PLACE dropout, we also investigate its efficacy in improving the SOTA DG methods, including MMLD \cite{matsuura2020domain} and SagNet \cite{nam2021reducing}.
MMLD is an adversarial domain-invariant learning method that trains the domain-generalized model by extracting features to confuse the domain discriminator.
SagNet is a regularization method that decouples content and style features from the representations based on statistics and reduces the style bias of the model to enhance its generalization.
As presented in Tab.~\ref{tab: DG SOTA methods}, both the SOTA methods gain significant improvement on DeepAll$^{++}$, \textit{i.e.}, MMLD gets $3.25\%$ ($85.08\%$ vs. $81.83\%$) and SagNet gets $2.20\%$ ($85.15\%$ vs. $82.95\%$).
The results verify the effectiveness of our strong baseline.
Moreover, among these methods, PLACE dropout achieves the best performance, which proves its superiority and effectiveness to mitigate the overfitting issue of the model on source domains.

\end{document}